\documentclass[twocolumn,10pt]{article}

\usepackage[a4paper,margin=1in]{geometry}
\usepackage{times}
\usepackage{graphicx}
\usepackage{amsmath,amssymb}
\usepackage{hyperref}
\usepackage{titlesec}

\usepackage[utf8]{inputenc}
\usepackage{amsmath,amssymb,amsfonts}
\usepackage{graphicx}
\usepackage{booktabs}
\usepackage{geometry}
\usepackage{comment}
\usepackage[normalem]{ulem}
\usepackage{tikzducks}
\usepackage{subcaption}
\usepackage{bm}
\usepackage{multirow}
\usepackage{amsthm}
\usepackage{mathrsfs}
\usepackage[title]{appendix}
\usepackage{xcolor}
\usepackage{textcomp}
\usepackage{manyfoot}
\usepackage{listings}
\usepackage{algorithm}
\usepackage{algpseudocode}
\usepackage{soul}
\usepackage{longtable}
\usepackage{pifont,xspace}
\usepackage[table]{xcolor}
\usepackage{tabularx}
\usepackage{makecell}
\usepackage{authblk}

\definecolor{lightGreen}{rgb}{0.30,0.56,0}

\definecolor{eladiocolor}{HTML}{20C9BB}

\definecolor{davidecolor}{HTML}{EF330B}

\newcommand{\avgRealTraffic}{y}
\newcommand{\simulator}{\hat{y}}

\newcommand{\region}{\rho}
\newcommand{\regionSet}{\mathcal{R}}

\newcommand{\edgeSet}{\mathcal{E}}
\newcommand{\edgeSetSensors}{\bar{\mathcal{E}}}

\newcommand{\todayDate}{May 19, 2026}

\newcommand{\counts}{y}      
\newcommand{\pred}{\hat{y}}  
\newcommand{\regpaths}{\Phi}          
\newcommand{\regpath}{\phi}           
\newcommand{\linkpath}{\ell}          
\newcommand{\x}{x}           
\newcommand{\C}{C}           
\newcommand{\regions}{\mathcal{R}}    
\newcommand{\loss}{\mathcal{L}}       

\newcommand{\numStabilityVar}{\gamma}

\newcommand\copyrighttext{%
  \footnotesize {\centering\textbf{\color{red}(*) Preprint version. This manuscript is currently under review for possible publication in the ACM SIGSPATIAL 2026 Conference.
}}}
\newcommand\copyrightnotice{%
\begin{tikzpicture}[remember picture,overlay]
\node[anchor=south,yshift=10pt] at (current page.south) {\fbox{\parbox{\dimexpr\textwidth-\fboxsep-\fboxrule\relax}{\copyrighttext}}};
\end{tikzpicture}%
}

\title{Simulation-Free Estimation of Traffic Flows from Sparse Count Data*}

\date{}

\author[1,2]{Davide Andrea Guastella}
\author[2,3]{Gianluca Bontempi}

\affil[1]{\small Aix Marseille University, CNRS, LIS, Marseille, France}
\affil[2]{\small Université Libre de Bruxelles, Machine Learning Group, Brussels, Belgium}
\affil[3]{\small WEL Research Institute, Wavre, Belgium}

\affil[ ]{\texttt{davide.guastella@lis-lab.fr}}
\affil[ ]{\texttt{gianluca.bontempi@ulb.be}}

\begin{document}

\maketitle
\copyrightnotice
\begin{abstract}
We propose a method for estimating time-varying traffic flow patterns from sparse aggregated vehicle counts. The method partitions the study area into spatial regions, constructs a set of feasible region-to-region routes, and solves a weighted least-squares optimization problem to determine the number of vehicles to allocate on each route. A weighted contribution matrix encodes sensor coverage, steering the optimizer toward flow configurations that are directly observable by sensors. Edge-level trajectories are then derived by scoring candidate routes against the temporal and volumetric profiles of aggregated regional sensor counts.

The method is evaluated on the Brussels road network using real and synthetic traffic data. Results show that the proposed approach reproduces the daily traffic profile in the input data and outperforms the baseline methods at a fraction of the computational cost. 
The source code and two realistic traffic scenarios are publicly available\footnote{\textbf{NOTE: Code and models will be made available upon acceptance}}.
\end{abstract}

\section{Introduction}

Vehicular traffic is generally observed through a network of fixed sensors such as induction loops, cameras, or radar devices. These devices cover only a fraction of the road network and are subject to failures and maintenance gaps. The central challenge is therefore to estimate a traffic model from these incomplete observations to support the evaluation of urban policies, managing congestion, and reducing emissions. Herein, estimating a traffic model refers to constructing a set of routes, each associated with a vehicle and a departure time, that, when provided as input to a traffic simulation software package, reproduces the observed traffic counts at the corresponding real-world locations.

Simulation-based methods iteratively adjust a traffic model until the simulated counts match the observed ones, achieving high accuracy but at significant computational cost: each evaluation requires a full traffic simulation, making them impractical for large networks or operational settings. Offline methods are faster but typically require a prior OD matrix (depicting the flow of passengers between various locations within a network~\cite{GALLIANI2024104246}), GPS traces, or dense sensor coverage to constrain the solution. A largely overlooked aspect of calibration methods is the difficulty of fine-grained adjustment at the edge level. Direct global OD estimation approaches optimize flows to match sensor counts on individual edges, making the problem highly detailed and difficult to stabilize. By instead operating through a hierarchical regional structure, the proposed method abstracts the calibration problem to the level of spatial aggregates, yielding a simpler approximation while ensuring regional concordance by construction.

In this article, we address a key challenge: \emph{can accurate, time-varying traffic flow patterns be estimated from sparse sensor counts alone, without any feedback from traffic simulation and without a prior OD matrix, at a computational cost that scales to city-level networks?}

\begin{figure}[!ht]
    \centering
    \includegraphics[width=\linewidth]{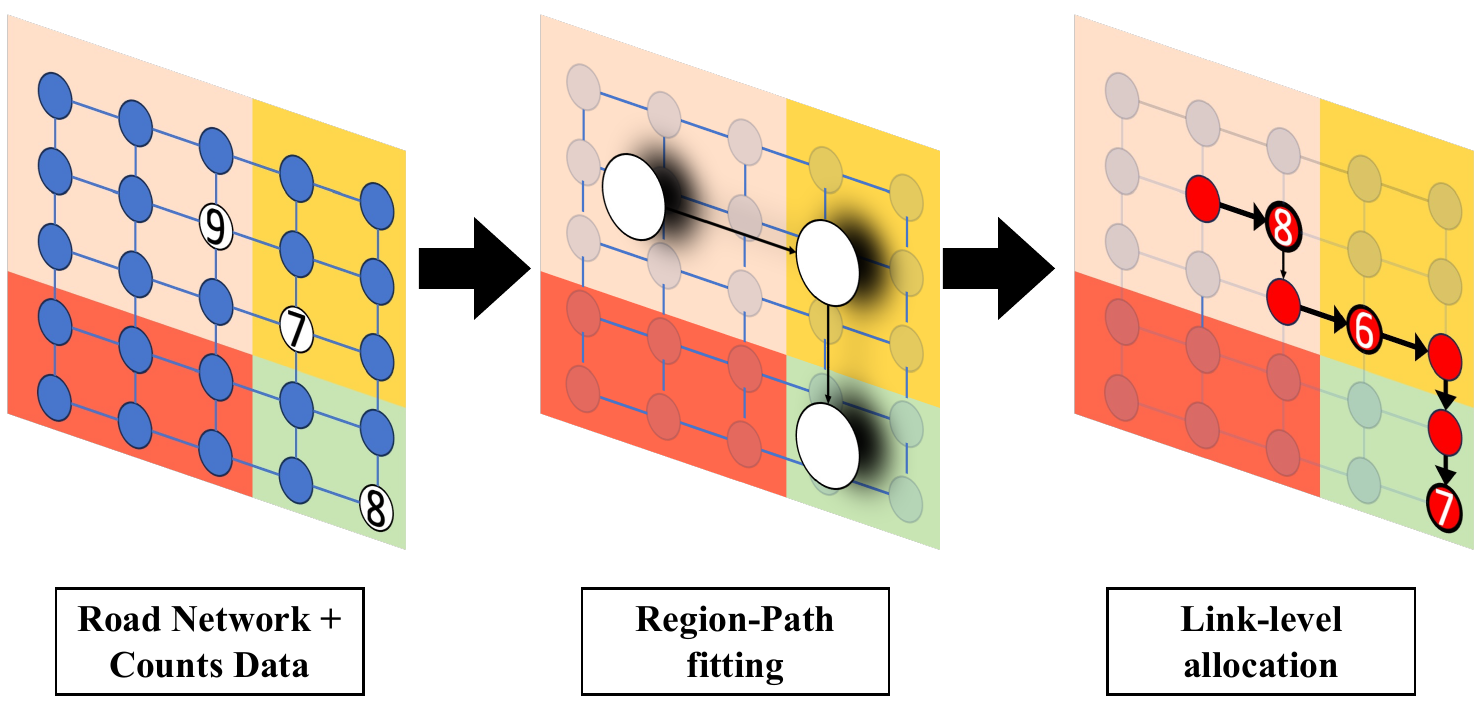}
    \caption{The proposed method operates hierarchically: the environment is first divided into arbitrary regions, then macroscopic region‑routes are fitted to the average traffic observed within each region, and finally vehicles are allocated to one or more edge‑level routes that are compatible with the assigned region route.}
    \label{fig:method}
\end{figure}

To achieve this, we propose a simulation-free framework for reconstructing time-varying edge-level traffic flows from sparse aggregated sensor counts through region-based flow assignment and route-consistent trajectory reconstruction. The proposed method does not require feedback from the simulator to adjust the traffic model and fit the input counts data. The method operates at two levels. At the regional level, the study area is partitioned into spatial regions and vehicle flows are assigned to feasible region-to-region routes by solving a weighted least-squares optimization problem. At the edge level, each regional route is refined into a concrete road-network trajectory by scoring candidate edge-level routes against the temporal and volumetric profiles of aggregate sensor counts. This hierarchical decomposition is the structural core of the proposed method and is illustrated in Figure~\ref{fig:method}. We use the open-source simulator SUMO to validate the results obtained by the proposed method.

The remainder of this paper is organized as follows: first, we discuss different state-of-the-art methods to address traffic calibration (Section~\ref{sec:background}). Next, we formalize the problem addressed in this paper (Section~\ref{sec:probstatement}). The following section introduces the proposed technique for calibrating traffic models using traffic count data (Section~\ref{sec:method}). We then illustrate the experimental setup and compare the results obtained from the proposed technique and two baseline methods (Section~\ref{sec:exp}). Finally, we summarize the main findings of this work and outline future research directions (Section~\ref{sec:conclusion}).

\section{Background Work}\label{sec:background}

Recent works on traffic calibration increasingly address the problem of reconstructing route flows or trajectories such that simulated traffic counts match observed measurements. This problem is intrinsically under-determined, as multiple trajectory configurations can reproduce the same aggregate counts, and different approaches resolve this ambiguity through simulation-based or optimization-based strategies.

Simulation-based calibration methods generate and iteratively refine demand parameters until the counts produced by the simulator align with ground-truth observations. Tang et al.~\cite{tang_parallel-computing-based_2024} propose parallelizing two standard heuristic optimizers, a genetic algorithm and particle swarm optimization for calibrating driver-behavior parameters of a microscopic simulator. At each iteration, a population of parameter sets is evaluated by running independent simulation replications, and the resulting simulated counts are compared against real measurements to guide the next generation of candidates; parallelizing these replications across cores reduces the calibration time from several hours to under one hour. Daguano et al.~\cite{daguano_automatic_2023} replace the repeated simulation calls with a Multi-Layer Perceptron trained to approximate the input-output behavior of the simulator, and use this surrogate to search for parameter configurations that reproduce observed traffic patterns. Osorio~\cite{OSORIO201918} proposes a framework that incorporates a closed-form analytical traffic model as a surrogate of the simulator within a simulation-based optimization procedure. The analytical model provides approximations of traffic performance measures, allowing the optimization algorithm to explore the OD search space efficiently while limiting the number of simulator evaluations. This hybrid strategy combines computational efficiency with the behavioral realism of detailed traffic simulation, including congestion propagation and route dynamics. Nevertheless, because the simulator must still remain in the optimization loop for iterative calibration and validation, the approach can lead to significant computational cost.

To limit computational cost, Roocroft et al.~\cite{roocroft_flow_2025} propose a simulation-free framework that constructs static traffic assignment models from loop-detector counts. The road network is first simplified by partitioning nodes into \emph{communities} using a modularity maximization algorithm from graph theory; this enables reducing the search space in the OD estimation problem. The approach scales to large national road networks and avoids repeated simulation runs. However, it relies on static assignment assumptions (Wardrop user equilibrium with standard volume-delay functions), and therefore does not model congestion dynamics or adaptive route choice, which limits its fidelity under varying traffic conditions.

Within this simulation-free perspective, several works focus on estimating route or OD flows under count constraints, often combining traffic flow models with additional data sources. Englezou et al.~\cite{ENGLEZOU2024213} introduce an OD matrix estimation framework that combines disaggregated loop-detector measurements with a macroscopic traffic flow model. Their approach is built on the route-based Cell Transmission Model (CTM), a discretized representation of traffic flow where each road link is divided into cells that propagate vehicle densities and flows according to conservation laws and fundamental diagram parameters. The authors first formulate the estimation task as a high-dimensional convex least-squares problem (DODLS), and then derive a reduced quadratic program (DODQP) that preserves linearity while lowering computational cost. Using the Nguyen--Dupuis benchmark network~\cite{nguyen_net}, they show that DODQP achieves nearly identical accuracy to DODLS (e.g., RMSE $\approx 2$~veh/h at full sensor coverage) but is up to sixty times faster. The method does not require simulation-in-the-loop; however, it remains tied to free-flow operating conditions and requires significant sensor coverage to reduce under-determination.

Cao et al.~\cite{CAO2024104850} propose a data-driven, time-dependent OD estimation framework built on a joint origin-destination-path-choice formulation. Shortest-path sets are first computed from static and dynamic link attributes (road type, travel time); the resulting path-share fractions and production-attraction totals are combined to derive a prior OD matrix, which is then scaled and constrained to be consistent with observed link flows. Because the resulting system of equations is often rank-deficient in large networks, Principal Component Analysis is applied to reduce the dimensionality of the problem before solving, and the solution is projected back to the full OD space. The method is evaluated on seven real-world road networks and outperforms state-of-the-art baselines; however, the joint path-choice formulation grows rapidly in size with network scale, making it challenging to apply to very large or sensor-sparse settings.

Wei et al.~\cite{WEI2025668} propose a method for estimating both OD flows and link flows using GPS probe trajectories. The method constructs route sets from empirical travel times, calibrates route choice using a Logit model, and solves a weighted non-negative least-squares problem that fits estimated route flows to link counts, with adherence to a prior OD matrix. Experiments on the Stockholm network show that the method achieves $R^2$ values around $0.90$ on the training sensors and $0.54$ on the held-out test sensors, with RMSE reductions of more than 60\% compared to assigning the prior OD matrix directly. The dependence on GPS probe data and a prior OD matrix limits applicability in sparse-data scenarios.

Overall, these approaches highlight a common limitation: while they avoid simulation-in-the-loop and improve computational efficiency, they either rely on strong modeling assumptions (e.g., free-flow conditions or predefined route choice models), require additional data sources such as prior OD matrices or trajectory data, or face scalability issues as network size increases. This motivates the development of methods that directly estimate trajectory flows from sparse count data without requiring simulation feedback, while remaining computationally efficient.

Compared to existing approaches, we introduce a method that estimates traffic models from raw sensor data without the need for simulation feedback. By formulating the problem as an optimization over region-level aggregated counts, the approach is fast and scalable to city-level networks, and does not require any prior OD demand matrix. Unlike recent large-scale OD estimation approaches (e.g., Zhang et al.~\cite{zhang2025osorio}), our method directly infers route-level traffic flows from sparse count data without assuming any traffic flow model or requiring simulation or travel time information. Its benefits are as follows:

\begin{itemize}
    \item \textbf{No prior OD matrix is required}: the proposed method estimates traffic models based on sparse count data. Trajectories are estimated based on plausible routes on the road network graph.
    \item \textbf{No simulation-in-the-loop}: estimated traffic flows are computed by an optimization model, avoiding the computational burden of simulation software to evaluate and adjust the traffic model.
    \item \textbf{Fully differentiable:} the objective function is smooth, enabling the use of gradient-based optimizers.
    \item \textbf{Scalable:} the dimensionality of the problem depends on the number of region-routes, not on the number of network links.
\end{itemize}

\section{Problem Statement}\label{sec:probstatement}

Table~\ref{tab:symbols} summarizes the symbols associated with the main concepts used to introduce the proposed method.

\begin{table}[!ht]
\centering
\caption{Notation used.}
\label{tab:symbols}
\small
\setlength{\tabcolsep}{4pt} 
\begin{tabularx}{\columnwidth}{@{} l >{\raggedright\arraybackslash}X @{}}
\toprule
\textbf{Symbol} & \textbf{Meaning} \\
\midrule
$\edgeSet$ & Set of edges in the road network. \\
$\edgeSet(\region) \subset \edgeSet$ & Set of edges in the region $\region$. \\
$\edgeSetSensors$ & Set of sensored edges. \\
$\edgeSetSensors_\linkpath \subset \edgeSetSensors$ & Set of sensored edges within the edge-level path $\linkpath$. \\
$\edgeSetSensors_\regpath \subset \edgeSetSensors$ & Set of sensored edges within the region-level path $\regpath$. \\
$\regions=\{\region_1,\region_2,...,\region_n\}$ & Set of spatial regions. \\
$\regpaths=\{\regpath_1,\regpath_2,...,\regpath_k\}$ & Set of feasible region-to-region routes. \\
$\linkpath_i \in \regpath$ & $i$-th edge-level route compatible with the region route $\regpath$. \\
$\Pi(\regpath)$ & Set of all candidate edge-level routes for the regional route $\regpath$. \\
$\counts^t$ & Vector of observed average traffic counts during time interval $t$. The $i$-th entry is the average traffic count collected by sensors in the $i$-th region; size $|\regions|$. \\
$\pred^t$ & Vector of predicted average regional traffic counts during time interval $t$; size $|\regions|$. \\
$\counts^t(\region)\in\mathbb{N}$ & Average traffic count measured by all sensors located within region $\region$, calculated during interval $t$. \\
$\counts^t_{\linkpath\in\Pi(\regpath)}$ & Vector of observed traffic counts along the edge route $\linkpath$ during interval $t$, where $|\counts^t_\linkpath| = |\edgeSetSensors_\linkpath|$. Each value $\counts^t_{\linkpath}(e)$, $e\in\edgeSetSensors$ is the number of vehicles observed in the edge $e$ during time interval $t$.\\
$\counts^t_{\regpath\in\regpaths}$ & Vector of observed traffic counts along the regional route $\regpath$; contains average counts observed by sensors within each spatial region during $t$; cardinality $|\counts^t_\regpath| = |\edgeSetSensors_\regpath|$. Each value $\counts^t_{\regpath}(\region)$ is the average number of vehicles observed in the region $\region$ during time interval $t$.\\
$\C$ & Contribution matrix linking routes to regions. \\
$\x^t$ & Vector of decision variables: number of vehicles to allocate to each region-route $\regpath\in\regpaths$ at time interval $t$; size $|\x^t|=|\regpaths|$. \\
$\loss :\mathbb{R}^{|\regpaths|}\to\mathbb{R}_0+$ & Objective function to be minimised. \\
$\lambda \in [0,1]$ & Weight of the sensor-based regularization term. \\
$Q_t$ & Matrix of size $|\regionSet| \times |\regionSet|$, $Q_t(\region_{r_1},\region_{r_2})$ is the probability that vehicles pass from $\region_{r_1}$ to $\region_{r_2}$ during time interval $t$.\\
$f_t:\regionSet\times\regionSet\to\mathbb{N}$ & Function that returns the number of transitions between two regions at interval $t$.\\

$\psi(\regpath,t) \in \mathbb{R}$ & Log-likelihood of a region-level route $\regpath$ at interval $t$ \\
$Pr(\region_{r_k} \to \region_{r_{k+1}}, t) \in [0,1]$ & Probability that vehicles pass from $\region_{r_{k}}$ to $\region_{r_{k+1}}$ during time interval $t$ \\
\bottomrule
\end{tabularx}
\end{table}

Let $\counts^t$ denote the average observed counts over a set of regions $\regions$ during the time interval $t$. The vector $\counts^t$ has dimension $|\regions|$, thus $\counts^t(\region)$ is the average traffic count measured by all sensors situated within region $\region$, calculated during the time interval $t$. The two primary hyperparameters of the method are the number of regions $|\regionSet|$ and the number of feasible routes in $\regpaths$. The choice of $|\regionSet|$ determines the spatial resolution of the observation vector $\counts^t$.

Before formalizing the problem, we clarify the two levels of representation used throughout this paper. An \emph{edge-level route} $\linkpath$ is an ordered sequence of edges \mbox{$(e_1,e_2,\cdots,e_m)\in\edgeSet$}. A \emph{region route} $\regpath$ is an ordered sequence of spatial regions \mbox{$(\region_1,\region_2,\cdots,\region_n)\in\regionSet$}. A region route is \emph{feasible} if every consecutive pair of regions is connected by at least one edge-level route. A region route is therefore an abstraction that groups together all edge-level routes sharing the same regional itinerary.

Let $\regpaths$ be the set of feasible region-to-region routes. The set $\regpaths$ is constructed exhaustively to include all feasible region paths between regions that satisfy a predefined minimum length constraint (expressed as the number of regions traversed). Therefore, the cardinality of $\regpaths$ affects the size of $\C$, and consequently the performance of the fitting method. Each route contributes to the traffic measured in the regions it traverses, represented through a contribution matrix $\C$. A contribution matrix $\C \in \mathbb{R}^{|\regpaths| \times |\mathcal{R}|}$ encodes how each route contributes to the regional traffic totals. Rather than a binary encoding, each entry is weighted by the sensor coverage of the route within the region:

\begin{equation}
    \C(\regpath,\region) = \frac{|\edgeSet_{\phi} \cap \edgeSetSensors(\region)|}{|\edgeSetSensors(\region)| + \numStabilityVar},
    \label{eq:weighted_C}
\end{equation}

\noindent where $\regpath \in \regpaths$, $\region\in \regionSet$, $\edgeSet_{\phi}$ is the set of edges traversed by regional route $\phi$, $\edgeSetSensors(\region) \subseteq \edgeSetSensors$ is the subset of sensor-equipped edges in region $\region$, and $\numStabilityVar$ is a small constant for numerical stability. A value of $\C(\regpath,\region) = 1$ indicates that route $\regpath$ passes through all sensors in region $\region$, while $\C(\regpath,\region) = 0$ indicates no sensor coverage. This weighting guides the optimization toward flow configurations that are directly observable from the available sensors.

For each region route $\regpath_i \in \regpaths$, we define the \emph{route flow} $\x^t (i)\geq0$ as the number of vehicles assigned to $\regpath_i$ during time interval $t$. Collecting all route flows into a vector $\x^t \in \mathbb{R}^{|\regpaths|}$, the predicted regional count vector is  $\pred^t = \C^\top \x^t$, where $\C$ is the contribution matrix defined in Equation~\ref{eq:weighted_C}. The estimation task is formulated as finding non-negative route flows that best reproduce the observed measurements, expressed as the following optimization problem:

\begin{equation}
    \x^{t} = \arg \min_{\x \geq 0}\,\mathcal{L}(\x)
    \label{eq:opt}
\end{equation}

\noindent where $\mathcal{L}$ evaluates the goodness of fit of the solution vector $\x^t$. The objective $\loss:\mathbb{R}^{|\regpaths|}\to\mathbb{R}_0+$ is a differentiable, non-negative function that measures the discrepancy between the predicted counts $\pred^t = \C^\top \x^t$ and the observed counts $\counts^t$:

\begin{equation}
  \mathcal{L}(\x^t) =
    (1 - \lambda)\,\mathcal{L}_{\mathrm{reg}}(\x^t)
    + \lambda\,\mathcal{L}_{\mathrm{aggr}}(\x^t),
  \qquad \lambda \in [0,1],
  \label{eq:loss}
\end{equation}

\noindent where $\mathcal{L}_{\mathrm{reg}}$ which penalizes per-region discrepancies, $\mathcal{L}_{\mathrm{aggr}}$ penalizes the relative deviation of the total predicted count from the total observed count, $\lambda$ controls the relative importance of the two terms. The full derivation of both terms is given in Section~\ref{sec:method}.

The goal is to solve this optimization problem to recover flow patterns that accurately reconstruct the input traffic counts $\counts^t$.

\section{Proposed Method}\label{sec:method}

The proposed method follows a hierarchical (from neighborhood scale to edge-level scale), simulation-free approach to reconstruct urban traffic flows from sparse count data. The methodology is structured in two primary stages: (\textit{i}) a region-level assignment that solves a global optimization problem to determine vehicle distribution across a simplified regional graph, and (\textit{ii}) an edge-level assignment that maps these regional flows onto the modeled road network. 

\subsection{Region-Level Assignment}

The region-level traffic estimation proceeds in three steps to solve at each time interval the optimization problem formalized in Equation~\ref{eq:opt}:

\begin{enumerate}
    \item Construct a set of feasible region-to-region routes based on the connectivity of the network.
    \item Represent each route as a contribution to the traffic observed in each region.
    \item Solve the optimization problem to assign to each regional route a number of vehicles such that the aggregated contributions match the observed counts, while penalizing routes that are unlikely to be detected by the available sensors.
\end{enumerate}

First, we divide time into discrete intervals (hours) and run the method separately for each. Then, we partition the study area into a set of regions $\regions$. Instead of reasoning directly at edge level, we assume that each trip follows a sequence of regions. A region delimits a part of the modeled environment, such as administrative boundaries, neighborhoods, or any spatial subdivision that groups together areas sharing similar characteristics, constraints, or functions. 

The estimation task is modeled as a weighted least-squares optimization problem (Equation~\ref{eq:loss}). Following, we describe the terms $\mathcal{L}_{\mathrm{reg}}$ and $\mathcal{L}_{\mathrm{aggr}}$. 

\paragraph{Weighted regional loss}
This term measures the per-region discrepancy between predicted and observed counts, weighted inversely by the observed count:

\begin{equation}
  \mathcal{L}_{\mathrm{reg}}(\x^t)
  = \frac{1}{|\mathcal{R}|}
    \sum_{\region \in \mathcal{R}}
    \frac{\left(\pred^t(\region) - \counts^t(\region)\right)^2}
         {\counts^t(\region) + \numStabilityVar},
  \label{eq:loss_reg}
\end{equation}

\noindent where $\numStabilityVar > 0$, with $\numStabilityVar \in \mathbb{R}$ is a constant for numerical stability. The weight $1/\left(y^t(\region) + \numStabilityVar\right)$ assigns greater importance to regions with low observed counts, preventing high-volume regions from dominating the gradient.

\paragraph{Aggregated loss}
Multiple flow configurations can reproduce the same distribution of region counts while yielding different total traffic volumes. For this reason, we introduce a further term that penalizes the relative discrepancy between the total predicted and observed volumes:

\begin{equation}
  \mathcal{L}_{\mathrm{aggr}}(\x^t)
  = \left(
      \frac{
        \sum_{\region \in \mathcal{R}} \pred^t(\region)
        -
        \sum_{\region \in \mathcal{R}} \counts^t(\region)
      }{
        \sum_{\region \in \mathcal{R}} \counts^t(\region) + \numStabilityVar
      }
    \right)^{\!2}.
  \label{eq:loss_vol}
\end{equation}

\noindent where $\numStabilityVar > 0$, with $\numStabilityVar \in \mathbb{R}$ is a constant for numerical stability. 

Both terms are dimensionless: $\mathcal{L}_{\mathrm{reg}}$ is a mean relative squared error and $\mathcal{L}_{\mathrm{aggr}}$ is a squared relative deviation. This common scale ensures that $\lambda$ controls a genuine tradeoff between local (region-level) accuracy and global traffic volume, independently of the magnitude of the observed counts or the number of regions.

\paragraph{Optimization}
The problem is solved using the Adam gradient-based optimizer~\cite{ZHANG2026129002}. At each iteration $i$, the decision variables are updated as:

\begin{equation}
  \x^{(i+1),t}
  = \mathrm{Adam}\!\left(
      \x^{(i),t},\,
      \nabla_{\x^t}\,\mathcal{L}\!\left(\x^{(i),t}\right)
    \right),
  \label{eq:update}
\end{equation}

\noindent with gradient clipping applied at each step to prevent unstable updates:

\begin{equation}
  \nabla_{\x^t}\,\mathcal{L}
  \;\leftarrow\;
  \frac{\nabla_{\x^t}\,\mathcal{L}}
       {\max\!\left(1,\;
         \tfrac{\|\nabla_{\x^t}\,\mathcal{L}\|_2}
               {\delta}
       \right)},
  \label{eq:clip}
\end{equation}

\noindent where $\delta = 10$ is the maximum gradient norm. Non-negativity of $\x$ is enforced by a softplus transformation, $x_i = \log(1 + e^{\tilde{\x}_i})$, so that the unconstrained variable $\tilde{\x}_i \in \mathbb{R}$ is optimized in place of $\x_i \geq 0$.

\subsection{Link-Level Assignment}

Once the number of vehicles assigned to each regional route has been calculated, we refine every regional sequence into a edge-level trajectory. Let $\regpath \in \regpaths$ be a region route of $n < |\regionSet|$ regions. For each region we choose a representative \emph{via edge} among the sensor-equipped edges in $\edgeSetSensors(\region)$ with probability $Pr(e)$ drawn from a uniform distribution. Formally, $Pr(e)=\frac{1}{|\edgeSet(\region)|}$, where $e\in\edgeSet(\region)$. The resulting sequence of edges $(e_1, e_2, \dots, e_n)$, with $n \leq |\regionSet|$, defines a chain of routing problems. For each pair of consecutive via edges $(e_k, e_{k+1})$, with $0\leq k<|\regionSet|-1$, we compute the shortest route:

\begin{equation}
\pi_k = \arg\min_{\pi \in \Pi(e_k, e_{k+1})} \mathrm{cost}(\pi),
\end{equation}

\noindent where $\Pi(e_k, e_{k+1})$ is the set of feasible routes edge-level routes between edges $e_k$ and $e_{k+1}$, $\mathrm{cost}(\pi)$ indicates the length of the route $\pi$. The final trajectory $\linkpath$ is obtained by concatenating all local shortest routes:

\begin{equation}
\linkpath_i = \pi_1 \,\Vert\, \pi_2 \,\Vert\, \dots \,\Vert\, \pi_{n-1}.
\end{equation}

\noindent where $\linkpath_i$ is the $i$-th edge-level route for the regional route $\regpath \in \regpaths$, $n-1$ is the number of paths connecting all the $n$ via edges. We denote $\Pi(\regpath)$ as the set of all candidate edge-level routes for the regional route $\regpath$. The cardinality of the set $\Pi(\regpath)$ is bounded by a fixed value for computational efficiency.

We score each candidate edge route $\linkpath$ using two complementary criteria, volume coverage and temporal shape similarity, to assess how well it matches the traffic counts $\counts^t_\regpath$ on its parent regional route $\regpath$. Volume coverage measures the fraction of regional traffic captured by sensors along $\linkpath$ and is computed as follows:

\begin{equation}
  \mathrm{coverage}(\linkpath,\regpath, t)
    = \frac{\|\counts^t_{\linkpath}\|_1}{\|\counts^t_{\regpath}\|_1 + \numStabilityVar},
  \label{eq:coverage}
\end{equation}

\noindent where $\|\cdot\|_1 = \sum_t (\cdot)^t$ denotes the sum over all time intervals and $\numStabilityVar > 0$ ensures numerical stability. A value close to 1 indicates that the sensors in $\linkpath$ contribute significantly to the total traffic demand over the region path $\regpath$. The vector $\counts^t_\linkpath$ contains the observed traffic counts along the edge route $\linkpath$ during the time interval $t$, where $|\counts^t_\linkpath| = |\edgeSetSensors_\linkpath|$. The vector $\counts^t_{\regpath}$, with $|\counts^t_{\regpath}|=|\regionSet|$, contains the average observed traffic counts along the regional route $\regpath$.

The temporal shape similarity criterion measures how closely the temporal demand pattern of $\linkpath$ matches the aggregated traffic profile observed by the sensors in the regions of the parent region route. Both profiles are first converted to probability mass functions by L1 normalization:

\begin{equation}
  \tilde{\counts}^{t}_{\regpath}
    = \frac{\counts^t_{\regpath}}{\|\counts^t_{\regpath}\|_1 + \numStabilityVar},
  \qquad
  \tilde{\counts}^{t}_{\linkpath}
    = \frac{\counts^t_{\linkpath}}{\|\counts^t_{\linkpath}\|_1 + \numStabilityVar}.
  \label{eq:pmf_norm}
\end{equation}

\noindent where $\numStabilityVar > 0$, with $\numStabilityVar \in \mathbb{R}$ is a constant for numerical stability. The shape dissimilarity is then defined as the complement of the inner product between the two probability mass functions:

\begin{equation}
  \mathrm{shape}(\linkpath, \regpath,t)
  =1-(\tilde{\counts}^{t}_{\regpath})^{\top}\,\tilde{\counts}^{t}_{\linkpath}
  \label{eq:shape}
\end{equation}

We use the L1‑normalized inner product because it directly measures exact temporal overlap (the probability that two independently drawn time bins coincide), while noting that standard alternatives such as the Bhattacharyya coefficient~\cite{Bhattacharyya} or the Hellinger distance~\cite{hellinger} could also be employed. The overall score assigned to a candidate edge-level route $\linkpath$ is

\begin{equation}
  s(\linkpath, \regpath,t)
    = \bigl|1 - \mathrm{coverage}(\linkpath,\regpath,t)\bigr|
      + \mathrm{shape}(\linkpath,\regpath,t),
  \label{eq:link_score}
\end{equation}
\noindent where $\regpath$ is the region path that contains $\linkpath$, $t$ is the current time interval. Lower values indicate better agreement with both magnitude and temporal shape of the regional traffic distribution. 

Given a region route $\regpath$ with estimated flow $\x_\regpath$, the candidate edge-level routes are ranked by $s(\linkpath,\regpath,t)$ and converted to a probability distribution via a softmax over the negative scores:

\begin{equation}
    P(\linkpath \mid \phi,t)
= \frac{\exp\!\big(-s(\linkpath, \phi,t)\big)}
        {\sum_{\linkpath_i' \in \Pi(\phi)} \exp\!\big(-s(\linkpath', \phi,t)\big)}
\end{equation}

\noindent where $\Pi(\regpath)$ is the set of all candidate edge-level routes for the regional route $\regpath$, $\linkpath \in \regpath$ is a edge-level route. The expression $\linkpath \mid \phi$ indicates that the probability is calculated considering the edge-level path $\linkpath$, which is chosen among all the link path that belongs to $\regpath$. Vehicles assigned to the regional route are then distributed across the candidate edge routes by sampling from this distribution. 

\section{Experimental Results}\label{sec:exp}

This section presents the results obtained by the proposed method. First, we present the mobility scenario used in our study, followed by the presentation of the simulation tool and evaluation metrics. We then describe the results obtained by the proposed method for estimating traffic models from raw counts data, followed by an evaluation on a synthetic scenario.

\subsection{Mobility Scenario}

\begin{figure}[!ht]
    \centering
    \includegraphics[width=\linewidth]{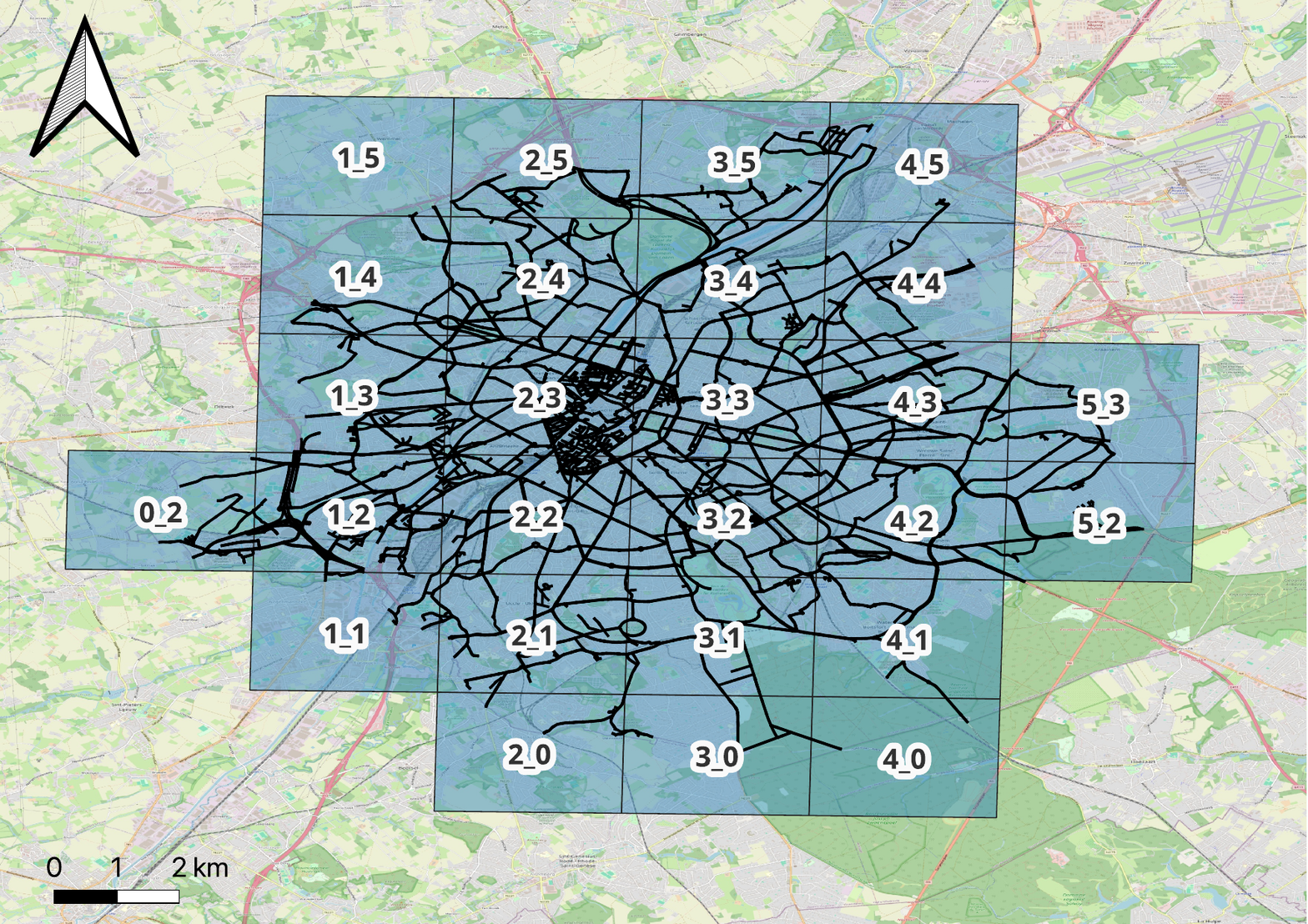}
    \caption{Brussels road network used as a case study. Each spatial region of 3000\,m$^2$ is associated to a unique identifier.}
    \label{fig:bxl_model}
\end{figure}

Figure~\ref{fig:bxl_model} shows the Brussels road network used in our experiments. This road network includes only roads where private vehicles are allowed to travel. Some edges are disconnected because we filter only vehicular roads, and due to the limited extent of the modeled physical environment. The presence of disconnected edges in the road network graph segments does not directly affect the calibration process: in the specific scenario we refer to, these parts do not include any sensors, and for this reason, the calibration algorithm does not insert any additional vehicles into these isolated segments.

For Brussels city scenario, the study area is divided into square regions of 3000\,m$^2$, and the dataset includes traffic counts collected the March 1, 2024, from 369 sensors, resulting in approximately 8800 samples.

We gather the necessary information concerning the topology for the road network from OpenStreetMap (OSM). The OSM files contain the definition of the road network, but also information on railways, subway lines, trams, public transportation stops, and other buildings information. We extract the information required to model the road network from the Geofabrik of the Belgium area\footnote{\url{https://download.geofabrik.de/europe/belgium.html}. Last visited: \todayDate}. For modeling the scenario of Brussels, we filter the geofabrik to include only the area of Brussels. We then convert the output in OSM format to SUMO, and filter the output road network to include only the vehicular road network. 

Traffic counts data has been downsampled by a factor 3. The downsampling factor depends on different factors such as the spatial extent of the modeled network, the density and type of roads included (in case filters are applied to retain only vehicular links). In practice, the goal is to ensure that the network can accommodate a realistic number of vehicles without saturating links or generating unstable traffic dynamics. In the case of the Brussels scenario, the road network is relatively large and includes only roads accessible to vehicles. This filtering step removes many minor links and can reduce the effective capacity of the modeled network. Downsampling the sensor data helps maintain a balance between realism and computational stability: it avoids injecting an excessive number of vehicles into a network whose topology has been simplified. Moreover, any inaccuracies introduced during the conversion from OSM to SUMO (missing connections, misinterpreted junction geometries) can amplify congestion artifacts.

The downsample value of 3 represents the maximum reduction that still maintains coherent traffic dynamics; stronger downscaling leads to excessive congestion, which in turn prevents the calibration algorithm from converging.

\subsection{Simulation Tool}

We use the open-source traffic simulator SUMO. SUMO can be used for microscopic simulation, where each vehicle and its dynamics are modeled individually, and mesoscopic simulation, where the movements of vehicles are modeled with queues and the traffic at intersections is modeled using a coarse model. We configure SUMO with a simplified microscopic model. This is done by simulating partially the behavior of vehicles in the intersections. When simulating traffic without considering the intersections, vehicles are still subject to right-of-way rules (waiting at traffic lights and minor roads) but they will appear instantly on the other side of the intersection after passing the stop line. The vehicles cannot block the intersection, wait within the intersection for left turns nor collide on the intersection\footnote{\url{sumo.dlr.de/docs/Simulation/Intersections.html\#internal_links}. Last visited: \todayDate}. In this work, we model junctions according to a microscopic model if the target queue of vehicles is jammed (that is, the occupancy of the edge is higher than 40\%).

\subsection{Evaluation Metrics}

We evaluate the following metrics to assess the accuracy of the estimate traffic model.

\begin{itemize}
    \item{\textbf{Mean Absolute Error (MAE)}: the absolute difference between the number of vehicles in the augmented and the real data. This metric indicates the number of vehicles that are exceeding in the augmented dataset compared to reality. The MAE is calculated using the following formula:
    \begin{equation}
        \textnormal{MAE}(\pred^t,\counts^t) = \frac{1}{|\edgeSetSensors|} \sum_{i=1}^{|\edgeSetSensors|} \left| \counts^t(i)-\pred^t(i) \right|
    \end{equation}
    
    \noindent where $\edgeSetSensors$ is the set of sensored edges, and $\simulator$ and $\avgRealTraffic$ the simulated and the real traffic counts at interval $t$ respectively.
    }
    \item{\textbf{Root Mean Square Error (RMSE):} The square root of the sum of the squared differences between the predicted and observed traffic counts, divided by the number of observations:
    \begin{equation}
        \textnormal{RMSE}(\pred^t,\counts^t) = \sqrt{\frac{1}{|\edgeSetSensors|}\sum_{i=1}^{|\edgeSetSensors|}\left(\counts^t(i) - \pred^t(i)\right)^2}
    \end{equation}
    
    \noindent where $\edgeSetSensors$ is the set of sensored edges, and $\pred^t$ and $\avgRealTraffic^t$ the simulated and the real traffic counts at interval $t$ respectively.
    }

    \item{\textbf{Normalized Root Mean Square Error (NRMSE)}: the RMSE normalized by the variance of the real traffic volume:
    \begin{equation}
    \small
        \textnormal{NRMSE}(\pred^t,\counts^t) = \frac{\textnormal{RMSE}(\pred^t,\counts^t)}{\sigma_y^2} = \frac{\sqrt{\frac{1}{|\edgeSetSensors|}\sum_{i=1}^{|\edgeSetSensors|}\left(\counts^t(i) - \pred^t(i)\right)^2}}{\frac{1}{|\edgeSetSensors|}\sum_{i=1}^{|\edgeSetSensors|}(\counts^t(i) - \bar{\counts}^t)^2}
    \end{equation}
    
    \noindent where $\bar{\counts}^t$ is the average of all traffic counts collected at interval $t$, $\sigma_y^2$ is its variance, $\edgeSetSensors$ is the set of sensored edges, and $\pred^t$ and $\counts^t$ the simulated and the real traffic counts at interval $t$ respectively.
}
\end{itemize}

\subsection{Results}

This section validates the results of the proposed method. The goal is to assess how well the estimate traffic reproduce the traffic profile of the real data. All experiments have been carried out on a server machine equipped with AMD Epyc processor (4th generation), and Linux operating system (kernel version 5.14). We do not consider any particular optimization technique to perform parallel computation.

The experimental methodology is as follows. For each experiment, the optimization method fit the solution vector $\x^t$ of vehicles to allocate to each region-level route at time interval $t$. Then, we generate a fixed number of distinct edge-level routes for each region route, that is, $|\Pi(\regpath)|=10,~\forall \regpath \in \regpaths$. This set is chosen according to a probability distribution derived from its score (Equation~\ref{eq:link_score}). The value $|\Pi(\regpath)|=10$ represents a practical trade-off between solution diversity and computational cost: preliminary experiments showed that the variance of the resulting traffic profiles stabilizes beyond $|\Pi(\regpath)| = 5$, while the additional gain from $|\Pi(\regpath)| > 10$ replicates is marginal relative to the increased simulation time.  We perform different experiments each with one region-level routes allocation, and several edge-level allocations. We then choose the best edge-level allocation as the one that allows, after simulation, minimize the gap between simulated and real traffic counts.

\begin{figure}[!ht]
    \centering
    \includegraphics[width=\linewidth]{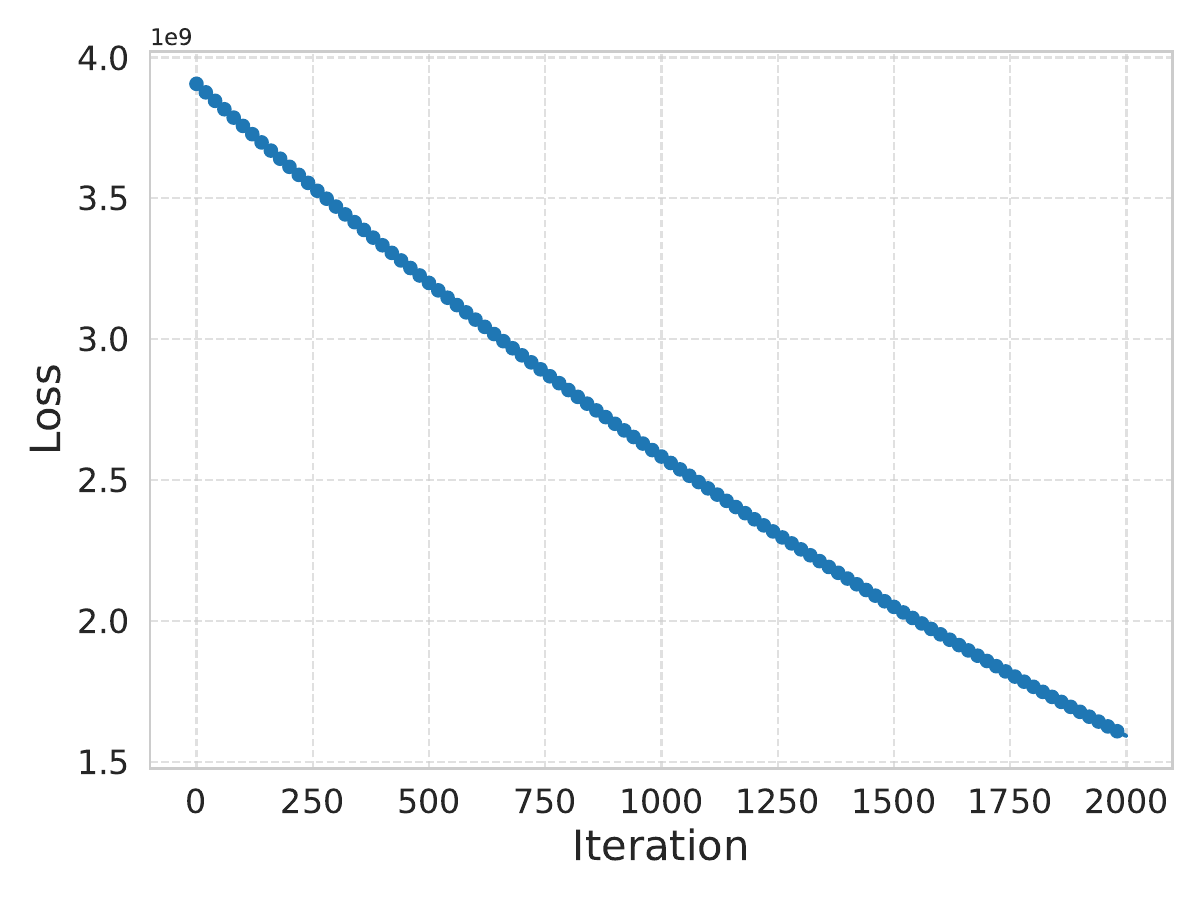}
    \caption{Profile of the average loss for the considered scenarios obtained by the proposed optimization method.}
    \label{fig:obj_fun}
\end{figure}

Figure~\ref{fig:obj_fun} shows the evolution of the objective function $\loss$ over the optimization process. The plot reports only the iterations in which the objective function achieved an actual improvement, omitting steps where no progress was observed.

Figure~\ref{fig:obj_fun} shows that the objective function doesn't reach a plateau, indicating that a higher number of iteration could help the optimizer to converge. Although the objective function $\mathcal{L}$ continues to decrease with additional iterations, we observed that beyond a certain threshold the estimated traffic profile begins to diverge from the real one, despite the loss remaining low. This behavior arises because the optimizer progressively concentrates vehicle flows on the subset of routes covered by sensors, since these are the only routes that directly reduce $\mathcal{L}_{\mathrm{reg}}$. As a result, unsensored edges receive progressively less traffic, and the total simulated volume falls below the observed one even though the per-region sensor counts are well reproduced.

\begin{figure}[!ht]
    \centering
    \includegraphics[width=\linewidth]{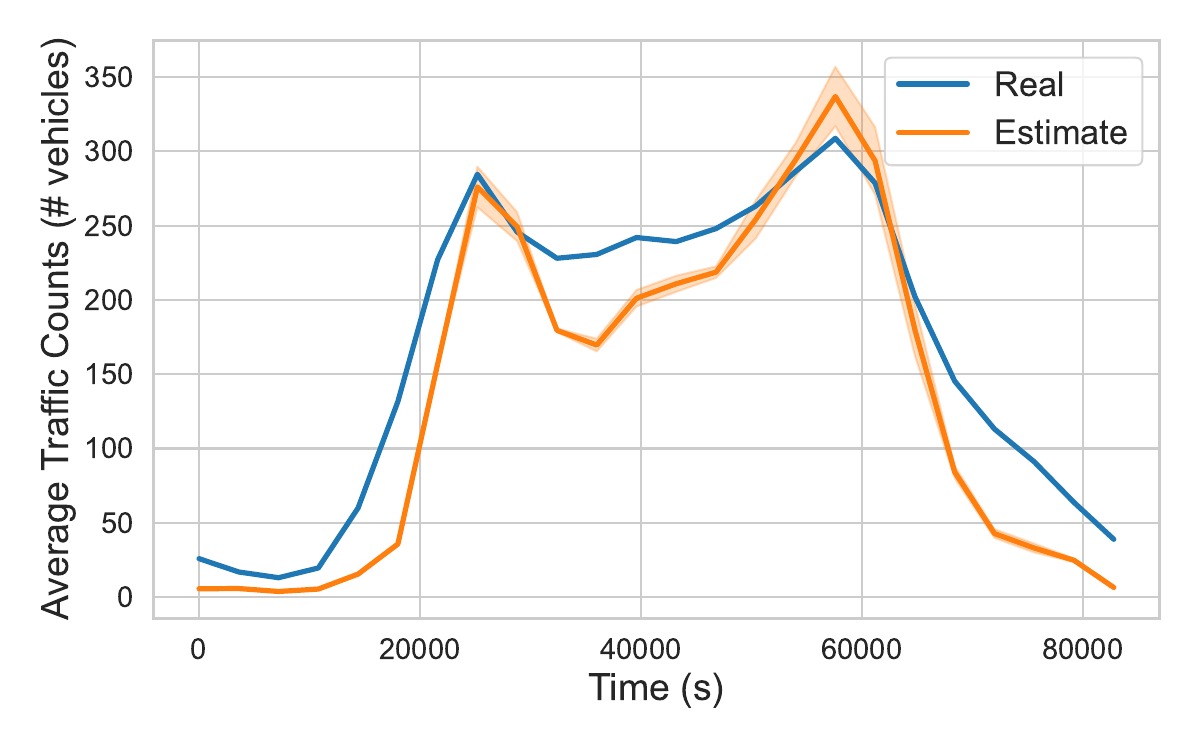}
    \caption{Average number of vehicles observed in all the regions and every time interval, in both the ground truth and the simulation.}
    \label{fig:traffic_vol}
\end{figure}

Figure~\ref{fig:traffic_vol} compares the average hourly traffic counts obtained by our proposal and the ground truth average traffic profile. This is obtained by calculating, for each time interval, the average among all the regions of the number of vehicles in the simulation and in the real data. This profile refers to the one obtained by simulating the best candidate edge-level routes allocation. 

Figure~\ref{fig:traffic_vol} shows the agreement between the total number of vehicles observed in the real data and the simulation. While some deviations appear during peak periods, where rapid fluctuations are harder to reproduce, the profile of the estimate traffic is consistent with the real trend. We highlight that the difference in the profile is low, indicating that the method is able to reconstruct the traffic profile pattern over time. The average standard deviation among all the experiment is of about 73 vehicles.

\begin{figure}[!ht]
    \centering
    \includegraphics[width=\linewidth]{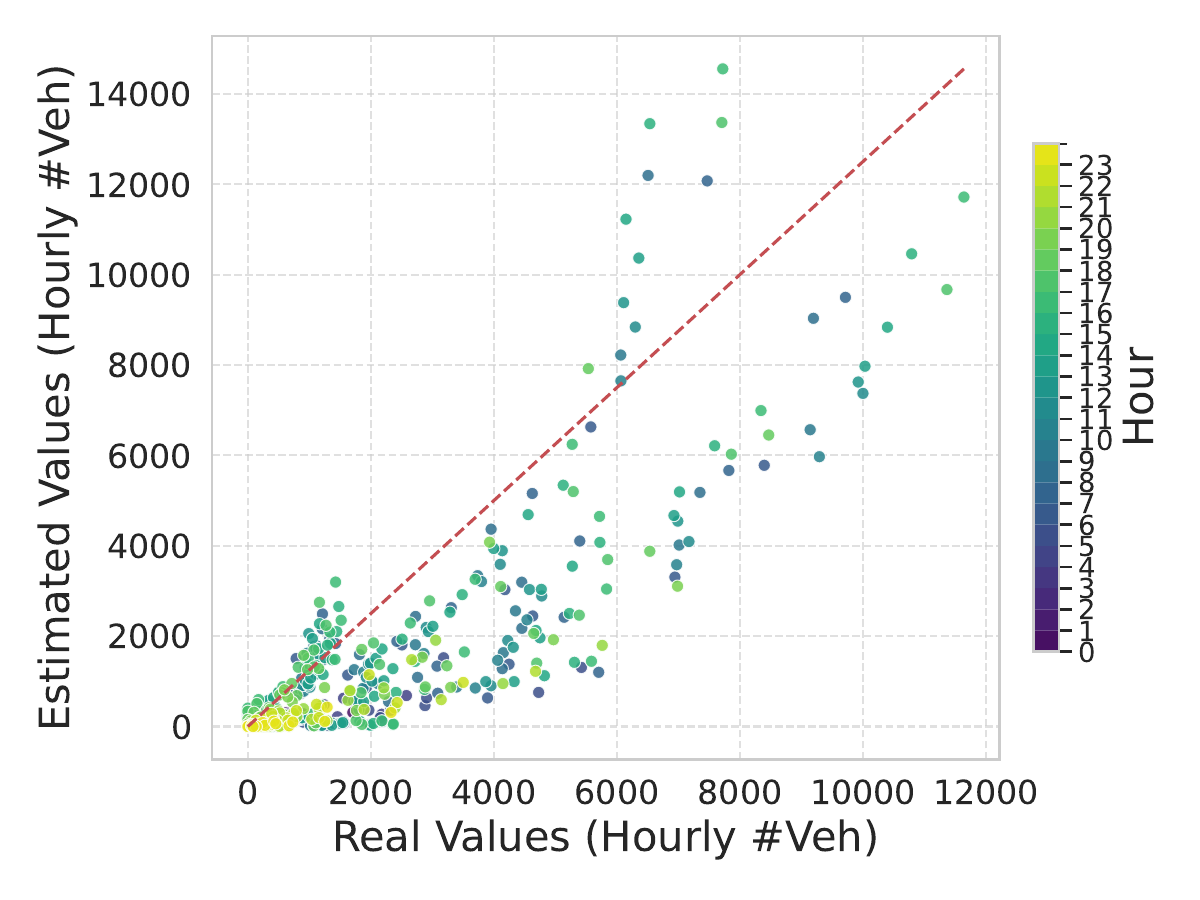}
    \caption{Total number of vehicles observed in all the regions and every time interval, in both the ground truth and the simulation.}
    \label{fig:scatter_plot_sensors}
\end{figure}

In the experiments we evaluated four values of $\lambda$ (used to regularize $\mathcal{L}_{\mathrm{aggr}}$ and $\mathcal{L}_{\mathrm{reg}}$): 0.2, 0.4, 0.6 and 0.8. Although the differences in the scatter plots of sensor counts (see Figure~\ref{fig:scatter_plot_sensors}) are not significant across the different $\lambda$ values, the overall estimated traffic profile is affected by this parameter. In the Brussels scenario, $\lambda=0.2$ contributes to a profile in which the peaks are similar between the real and estimated traffic.

\begin{figure}[!ht]
    \centering
    \includegraphics[width=\linewidth]{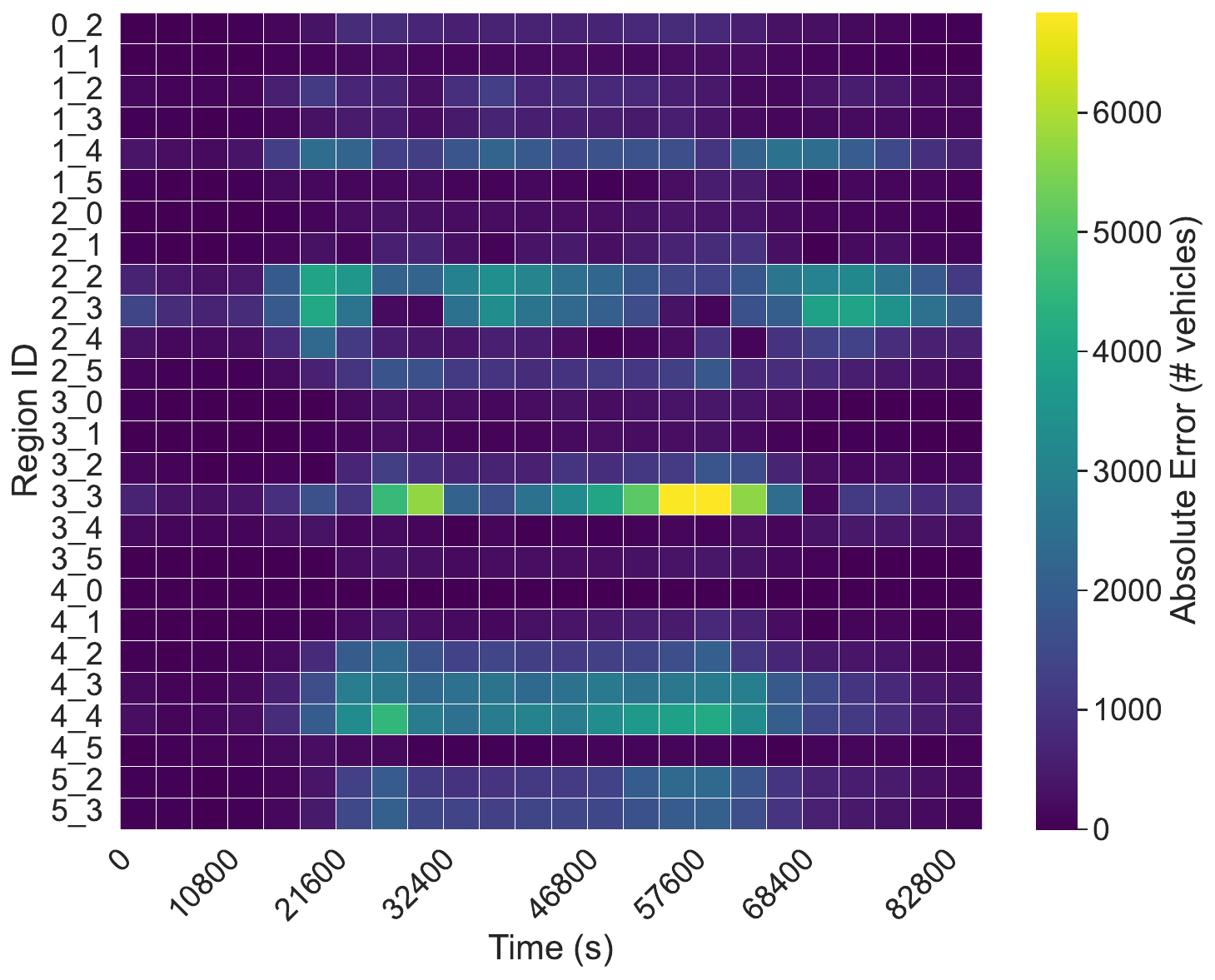}
    \caption{Absolute error (number of vehicles) per region and time interval.}
    \label{fig:err_reg_heatmap}
\end{figure}

Figure~\ref{fig:err_reg_heatmap} reports the regional error, defined as the absolute difference between estimated and real region counts for every time interval considered.

\begin{figure*}[!ht]
    \centering
    \begin{subfigure}[b]{0.49\textwidth}
        \centering
        \includegraphics[width=\linewidth]{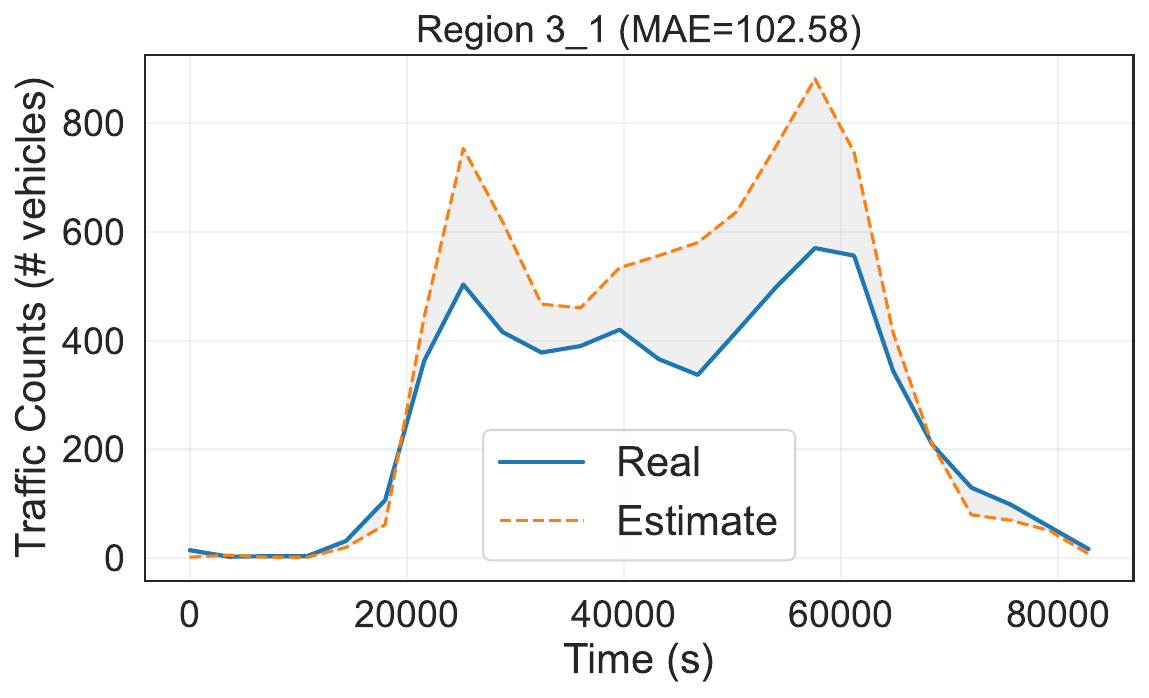}
        \caption{}
        \label{fig:reg_profile_a}
    \end{subfigure}
    \hfill
    \begin{subfigure}[b]{0.49\textwidth}
        \centering
        \includegraphics[width=\linewidth]{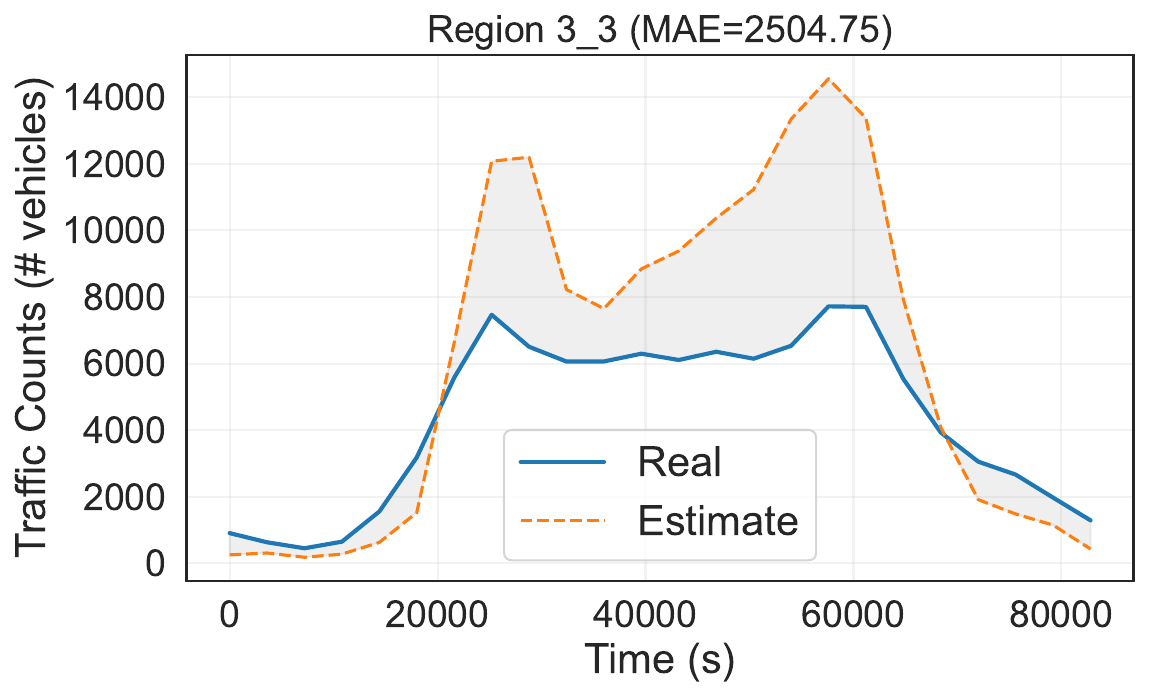}
        \caption{}
        \label{fig:reg_profile_b}
    \end{subfigure}    
    \caption{Average traffic counts for two spatial regions in the Brussels scenario. Figure~\ref{fig:reg_profile_a} approximatively matches the average real data in profile and magnitude, whereas Figure~\ref{fig:reg_profile_b} diverges in both shape and magnitude.}
    \label{fig:reg_profiles}
\end{figure*}

The error in Figure~\ref{fig:err_reg_heatmap} shows that the proposed method fits the average counts closely for most regions, meaning the simulated traffic profiles align well with the ground truth at the region level. However, a subset of regions exhibits larger absolute errors, which highlights two complementary issues: (\textit{i}) local ambiguity in how many distinct path combinations can produce the same region counts, and (\textit{ii}) the under‑determined nature of the inverse problem, that is, multiple path-level solutions can match aggregate counts while producing different local errors. As a result, the model can fit the global pattern well but still concentrate error in specific regions where path ambiguity or sparse observations make the reconstruction unstable. To illustrate this contrast, Figure~\ref{fig:reg_profiles} compares the traffic profiles for region 2\_3 (the worst region) and region 3\_1 (one of the best).

The profiles in Figure~\ref{fig:reg_profiles} indicate that the traffic pattern for each region follows the shape of the observed daily profile. The principal discrepancy lies in the magnitude: the estimated counts often differ from the observed vehicle numbers, producing in this case under‑estimation in some regions. Because the fitting procedure operates independently of the traffic simulator, it cannot predict simulator-specific effects such as congestion propagation; consequently, matching region‑level averages does not guarantee that simulated edge‑level dynamics will be accurate. Also, improving the fit for one region by adjusting path flows can degrade the solution elsewhere; this trade‑off reflects multiple path combinations that reproduce the same aggregates but distribute vehicles differently across the network.

Figure~\ref{fig:scatter_plot_sensors} compares the sensors-level real traffic counts with those produced by the simulation. This comparison provides a quantitative assessment of how accurately the estimated model reproduces the traffic volumes measured by real sensors.

The error in Figure~\ref{fig:scatter_plot_sensors} reflects the under-determined nature of the problem, as the choice of edge-level routes directly affects which sensors are activated in the simulation. Without running the simulator in advance, it is not possible to predict how vehicles will distribute across individual links once injected into a congested network setting, making sensor-level accuracy difficult to guarantee. Overall, the scatter confirms good accuracy at the regional level, but highlights that precise sensor-level reconstruction requires either a tighter coupling between the route choice model and the sensor layout, or an iterative simulation-based refinement that is outside the scope of the proposed simulation-free framework.

\begin{figure}[!ht]
    \centering
    \includegraphics[width=\linewidth]{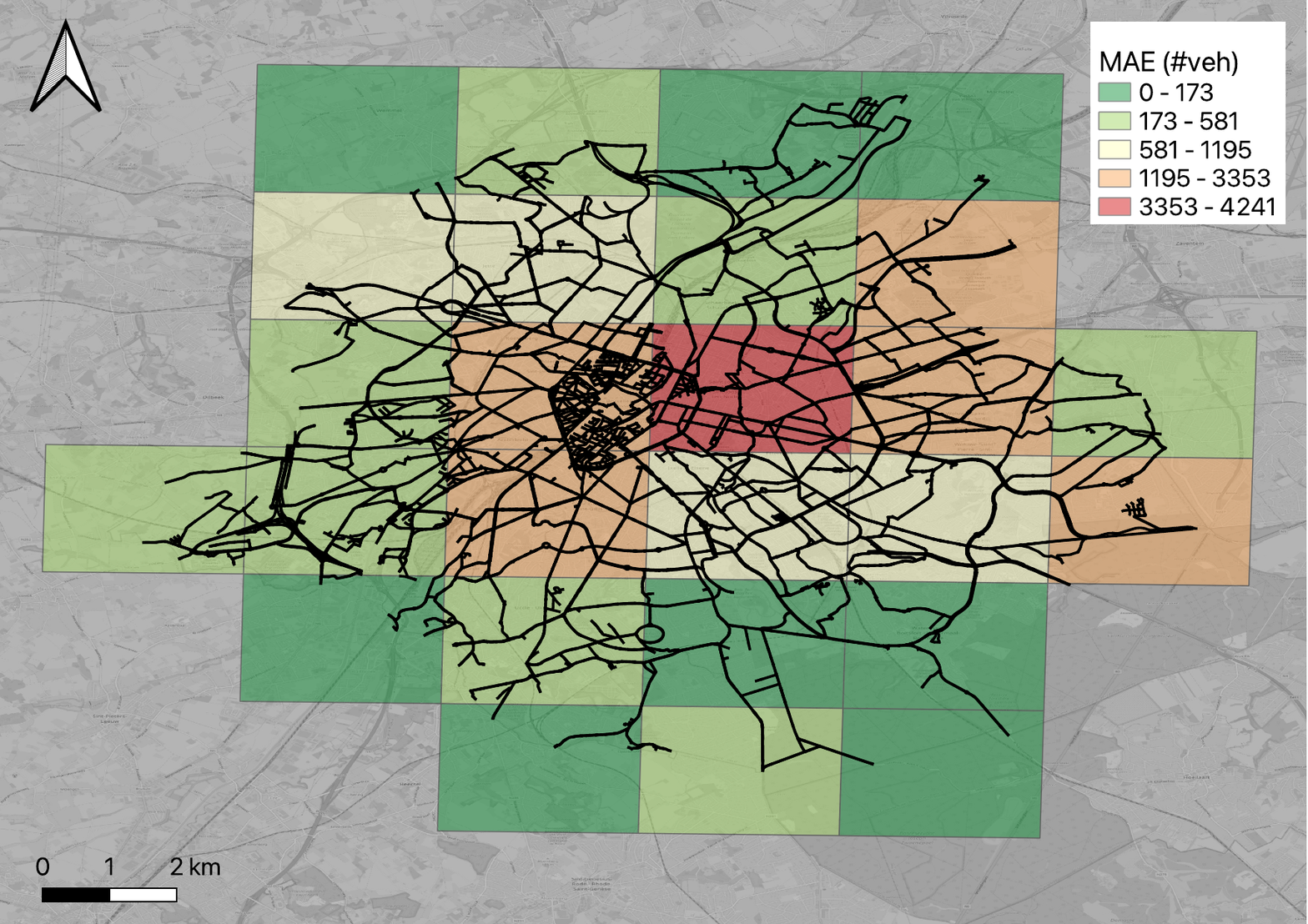}
    \caption{Spatial distribution of the traffic counts error (MAE) calculated over 24h.}
    \label{fig:mae_by_regions}
\end{figure}

Figure~\ref{fig:mae_by_regions} shows the spatial distribution of the 24-hour MAE across the Brussels network. While peripheral regions generally achieve low errors, the highest MAE (4.241) is concentrated in the central districts. This discrepancy is mainly due to the extreme traffic density and high saturation in the city center, where rapid demand fluctuations are harder to capture using aggregated regional counts. In these zones, the proposed method struggles to accurately distribute flow across the dense road network, leading to the localized spikes (the profile of the worst region is shown in Figure~\ref{fig:reg_profile_b}). 

\subsubsection{Comparison to Baseline}

We evaluate the proposed method against two baseline techniques: (\textit{i}) the calibration technique established in our previous research~\cite{guastella_calibration_2025}, and (\textit{ii}) the RouteSampler\footnote{\url{https://sumo.dlr.de/docs/Tools/Turns.html}. Last visited: \todayDate} tool available with the SUMO simulator. The simulation-based calibrator takes in input a set of traffic counts and corresponding sensor locations. It begins by creating a set of region-level routes (a region can be related to a neighborhood, or administrative boundaries), then it creates edge-level routes passing through the regions and through a set of sensors, using an exploration/exploitation tradeoff to avoid overfitting. The method is iterated until the simulation of the estimate model produces traffic counts that are close to the reals. The output is a set of vehicles, each defined by a starting time and an ordered sequence of edges representing its route. The environment is partitioned into a set of non-overlapping square regions. To ensure coherence with the discussed scenario, we use region sizes of 3000~m$^2$. We performed 10 different experiments using the simulation-based method.

RouteSampler takes as input a random traffic definition and filters the routes to match the real traffic density values~\cite{emode23}. Based on traffic counts information, this tool selects the routes from an initial set of trips (generated randomly using the \texttt{randomTrips}\footnote{\url{https://sumo.dlr.de/docs/Tools/Trip.html}. Last visited: \todayDate} tool available with SUMO) in such a way that the input traffic density values are matched to the output routes set~\cite{routesampler}. We used the default parameters for RouteSampler.

\begin{figure}[!ht]
    \centering
    \includegraphics[width=\linewidth]{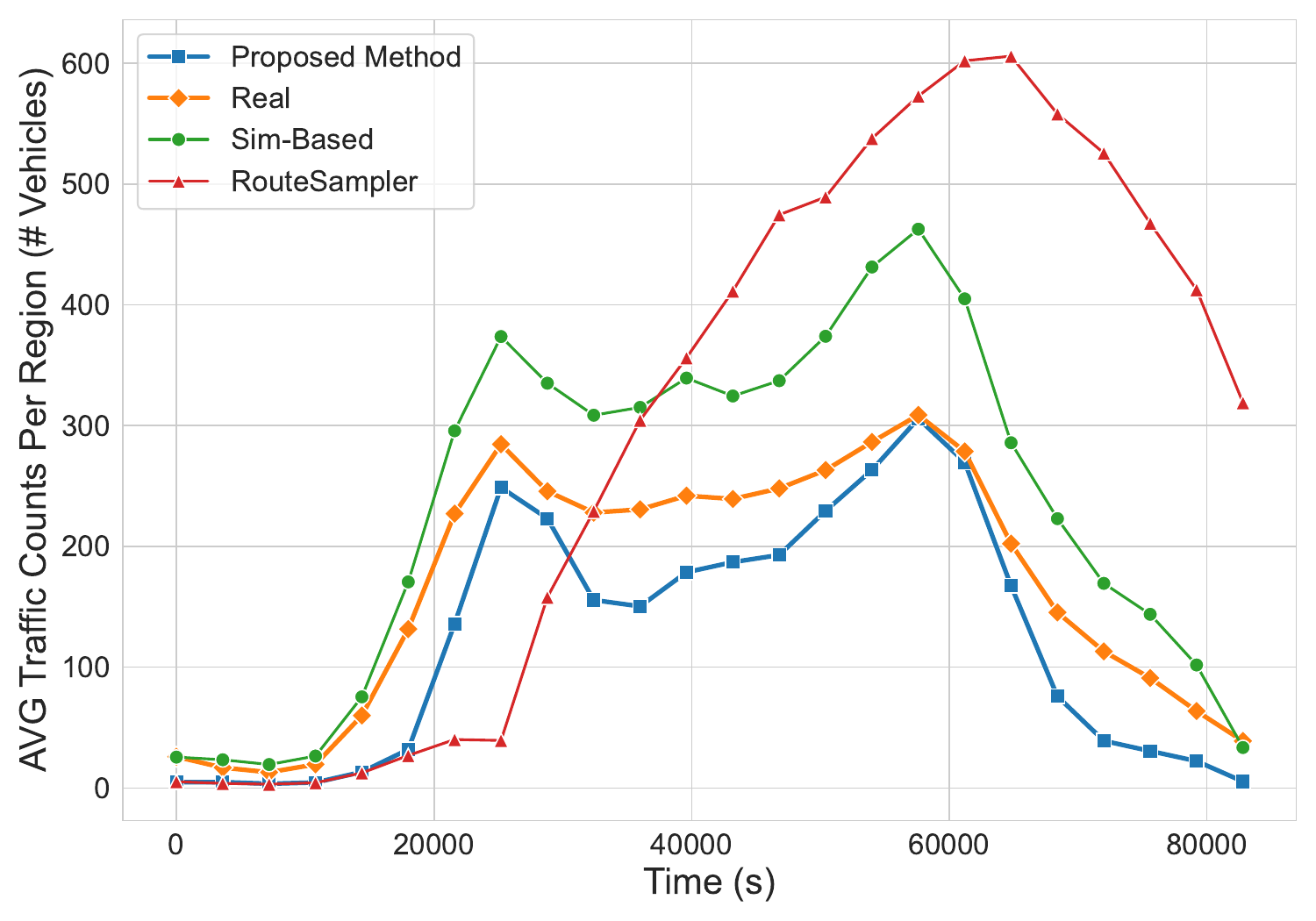}
    \caption{Comparison of the traffic profiles obtained by our proposal and the baseline methods. Results for RouteSampler have been downsampled by a factor 10 for visualization purpose.}
    \label{fig:all_avg_profiles}
\end{figure}

Figure~\ref{fig:all_avg_profiles} compares the average hourly traffic profiles produced by the proposed method, the simulation-based baseline, and RouteSampler against the real observations over the full 24-hour period. Note that the RouteSampler curve is displayed at one-tenth of its actual scale for visualization purposes, as its raw output substantially overestimates the real traffic volume.

The proposed method and the simulation-based baseline both reproduce the two-peak structure of the real profile with comparable accuracy in terms of average magnitude. RouteSampler fails to reproduce the traffic profile of the employed traffic data. 

Beyond profile accuracy, a key distinction between the proposed method and the simulation-based baseline lies in their generalization capability. The simulation-based approach requires running a traffic simulator at each calibration iteration to evaluate the current solution, making it computationally expensive and tightly coupled to the specific input dataset. The proposed method instead, estimates a parametric model of regional flows that can be stored, queried, and applied to new observations without re-running the full estimation pipeline. This property is particularly relevant in operational settings where traffic conditions change over time and the model must be updated or transferred to new time periods or sensor configurations at low computational cost. This is also remarkable when looking at Table~\ref{tbl:errors}, which compares the error obtained by our proposal and the baseline methods.

\begin{table}[!ht]
\caption{Error metrics for each calibration method compared to real data.}
\label{tbl:errors}
\centering
\begin{tabular}{lccc}
\toprule
\textbf{Method} & \textbf{MAE} & \textbf{RMSE} & \textbf{NRMSE} \\
\midrule
Proposed Method & 44.16 & 51.80 & 0.004 \\
Sim-Based~\cite{guastella_calibration_2025}       & 67.02 & 80.18 & 0.007 \\
RouteSampler~\cite{routesampler}    & 192.45 & 237.24 & 0.022 \\
\bottomrule
\end{tabular}
\end{table}

The errors in Table~\ref{tbl:errors} indicate that the proposed method outperforms both the simulation-based baseline and RouteSampler, achieving the lowest MAE (44.16) and RMSE (51.80). While the simulation-based method requires an average of 69 minutes to converge to a model that minimizes the distance to the real data, the proposed method requires only about 23 seconds to evaluate the region-level vehicle allocation over 24 hours, and just 0.2 seconds to evaluate a single edge-level route allocation. Each run of the SUMO simulator using the model calibrated using the proposed method requires about 98 seconds to perform a 24h simulation using a mesoscopic model. The routesampler pipeline instead requires over 3 hours.


\subsection{Synthetic Scenario Validation}\label{sec:synthetic}

We introduce a synthetic scenario to evaluate the ability of the proposed method to recover true mobility patterns under fully controlled conditions. We implemented a stochastic traffic generation procedure that distributes a fixed volume of vehicles over a 24-hour interval to replicate typical urban mobility patterns. Vehicle departures are allocated according to an hourly probability vector that incorporates a bimodal demand profile with two distinct peaks at 8:00 and 17:00, to ensure the synthetic data reflects realistic commuting behavior. We modeled a scenario where 80\,000 vehicles are distributed over all regions and time intervals, and fixed the minimum region route length to 2, 4, and 6. The calibration method has been configured with the matching minimum region route length.

Table~\ref{tab:region_performance} shows the error obtained from the proposed method using three synthetic scenarios, generated using region route length values of 2, 4, and 6.

\begin{table}[!ht]
\centering
\caption{Performance metrics of the proposed method using the synthetic scenarios, on the Brussels road network.}
\label{tab:region_performance}
\resizebox{\linewidth}{!}{
\begin{tabular}{ccccccc}
\toprule
\makecell{\textbf{Region Path}\\\textbf{Length}} &
\makecell{\textbf{Learning}\\\textbf{Rate}} &
\makecell{\textbf{Optimization}\\\textbf{Steps}} &
\textbf{RMSE} &
\textbf{MAE} &
\textbf{NRMSE} \\
\midrule
\textbf{2} & 0.01 & 100  & 65.86 & 57.48 & 0.59 \\
\textbf{4} & 0.01 & 100  & 75.80 & 66.38 & 0.81 \\
\textbf{6} & 0.01 & 2000 & 76.78 & 69.03 & 0.66 \\
\bottomrule
\end{tabular}
}
\end{table}

As discussed previously, the number of iterations affects directly the accuracy of the resulting model. Despite the convergence of the method with a high number of iterations, this leads to overfitting in sensor locations and compromise the overall model accuracy.

\subsubsection{Global Transition Matrix Recovery}

A transition matrix $Q_t$ at time interval $t$ is a row-stochastic matrix of size $|\regionSet| \times |\regionSet|$, where $\regionSet$ is the set of regions, and each entry

\begin{equation}
    Q_t(\region_{r_1}, \region_{r_2}) =
    \frac{f_t(\region_{r_1} \to \region_{r_2})}
         {\sum_{\region' \in \regionSet} f_t(\region_{r_1} \to \region')}
    \label{eq:tm}
\end{equation}

\noindent represents the probability that a vehicle currently in region $\region_{r_1}$ flows to region $\region_{r_2}$ during interval $t$; $f_t:\regionSet\times\regionSet\to\mathbb{N}$ returns the observed number of transitions between two spatial regions at time interval $t$.

\begin{figure}[!ht]
  \centering
  \includegraphics[width=\linewidth]{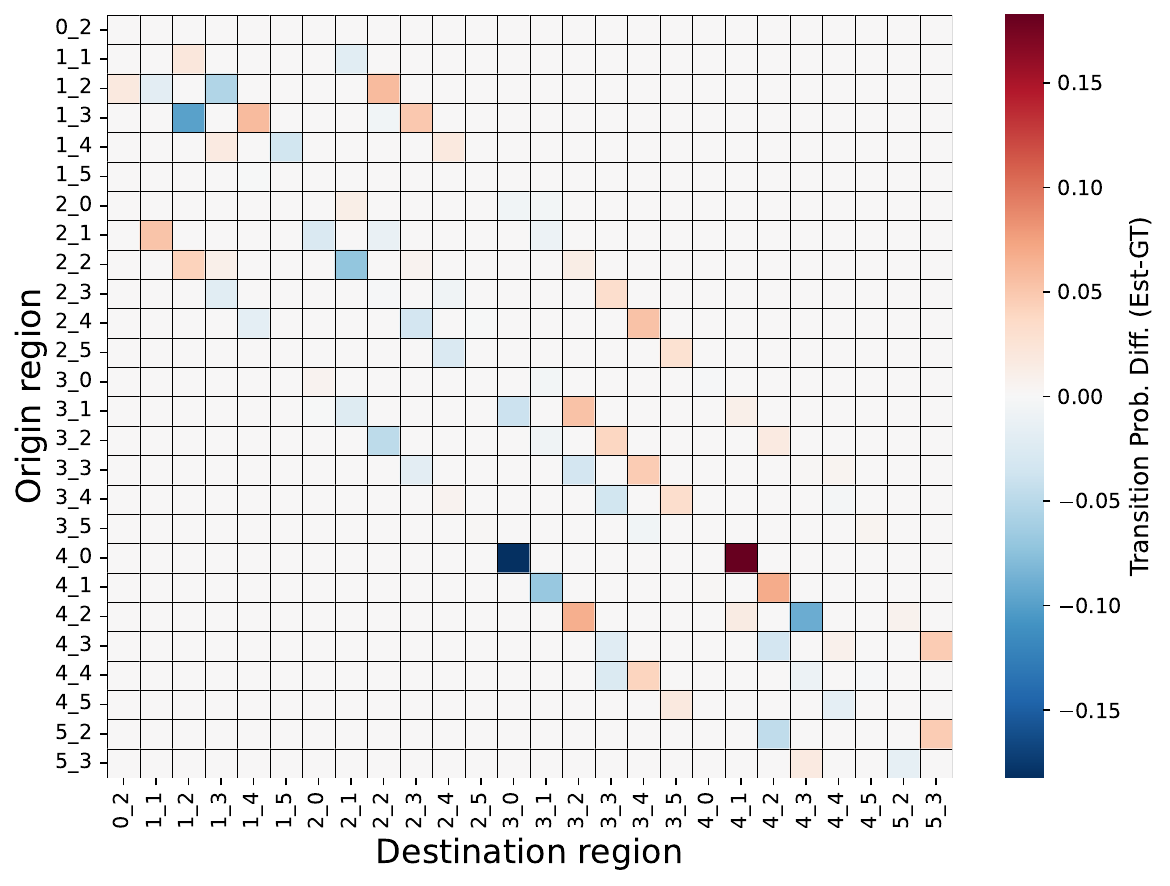}
  \caption{Element-wise difference $\hat{\bar{Q}} - \bar{Q}$ (red = overestimation, blue = underestimation). The dominant sparse block structure is preserved in the estimate. Residual errors are localized in a small number of OD pairs and do not exceed $\pm 0.20$ in probability.}
  \label{fig:tm_global}
\end{figure}

Figure~\ref{fig:tm_global} compares the aggregated (average over all time intervals) ground-truth transition matrix $Q$ (left) with the estimate transition matrix $\hat{\bar{Q}}$ (center), and their element-wise signed difference $\bar{Q} - Q$ (right).

The difference map reveals that residual errors are localized in a small number of off-diagonal entries, with the largest deviations reaching approximately $\pm 0.20$ in probability. These errors concentrated in a few OD pairs, indicating that the method occasionally misroutes flows between neighboring regions that share similar sensor coverage profiles. Instead, the majority of entries in the difference map are close to zero, proving that the estimated model preserves the global routing structure of the ground truth traffic.

\subsubsection{Routing Diversity}

\begin{figure}[!ht]
  \centering
  \includegraphics[width=.8\linewidth]{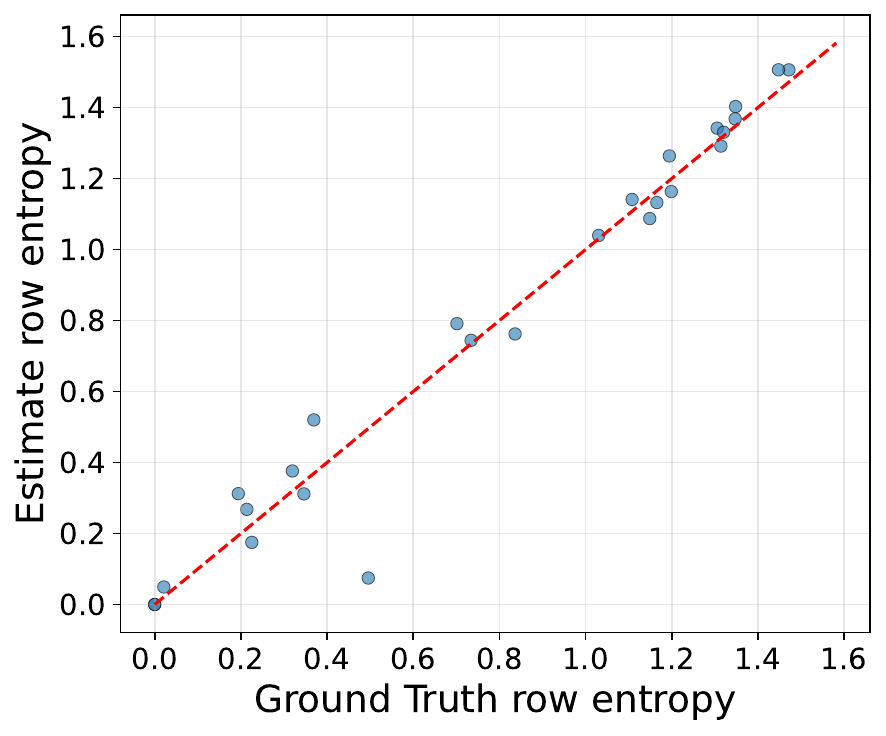}
  \caption{Per-region Shannon entropy of the ground-truth and estimated transition matrices. Each point corresponds to one origin region. Strong alignment along the identity line indicates that the estimated model reproduces the routing diversity of the ground truth model.}
  \label{fig:tm_structural}
\end{figure}

Figure~\ref{fig:tm_structural} compares the per-region Shannon entropy of the ground-truth and estimated transition matrices. For each origin region $\region$, the entropy \mbox{$H(\region) = -\sum_{\region'} p_{\region \to \region'} \log p_{\region \to \region'}$} quantifies the diversity of outgoing flows: low entropy indicates that vehicles are routed toward a single dominant destination region, while high entropy reflects genuine uncertainty across multiple destinations.

The scatter plot shows strong alignment along the identity line across the full range of entropy values ($[0,\,1.5]$ nats), indicating that the proposed method reproduces the routing diversity of each region. A small number of outliers fall below the identity line, indicating that the method slightly under-diversifies the outflow of a few regions, concentrating vehicles on fewer destinations than observed in the ground truth. This is consistent with the partial sensor coverage of the scenario: in regions where only a subset of corridors are instrumented, the method has limited information to distinguish between alternative routes and defaults to the most observable one.

\subsubsection{Trajectory Log-Likelihood}

The log-likelihood of a region-level route $\regpath$ at interval $t$ is defined as:

\begin{equation}
  \psi(\regpath,t) =
  \sum_{k=1}^{K-1}
  \log Pr\!\left(\region_{r_{k}} \to \region_{r_{k+1}},\, t\right),
  \label{eq:loglik}
\end{equation}

\noindent where $\region_{r_{k}},\region_{r_{k+1}} \in \regpath$, $r_k < r_{k+1}\leq |\regpath|$, $Pr(\region_{r_k} \to \region_{r_{k+1}}, t)$ is the probability that vehicles flow from $\region_{r_{k}}$ to $\region_{r_{k+1}}$ during time interval $t$ in the ground truth model. Higher (less negative) values of $\psi(\regpath)$ indicate that the route $\regpath$ contributes to transitions that are frequently observed in the ground truth model.

\begin{figure}[!ht]
  \centering
  \includegraphics[width=\linewidth]{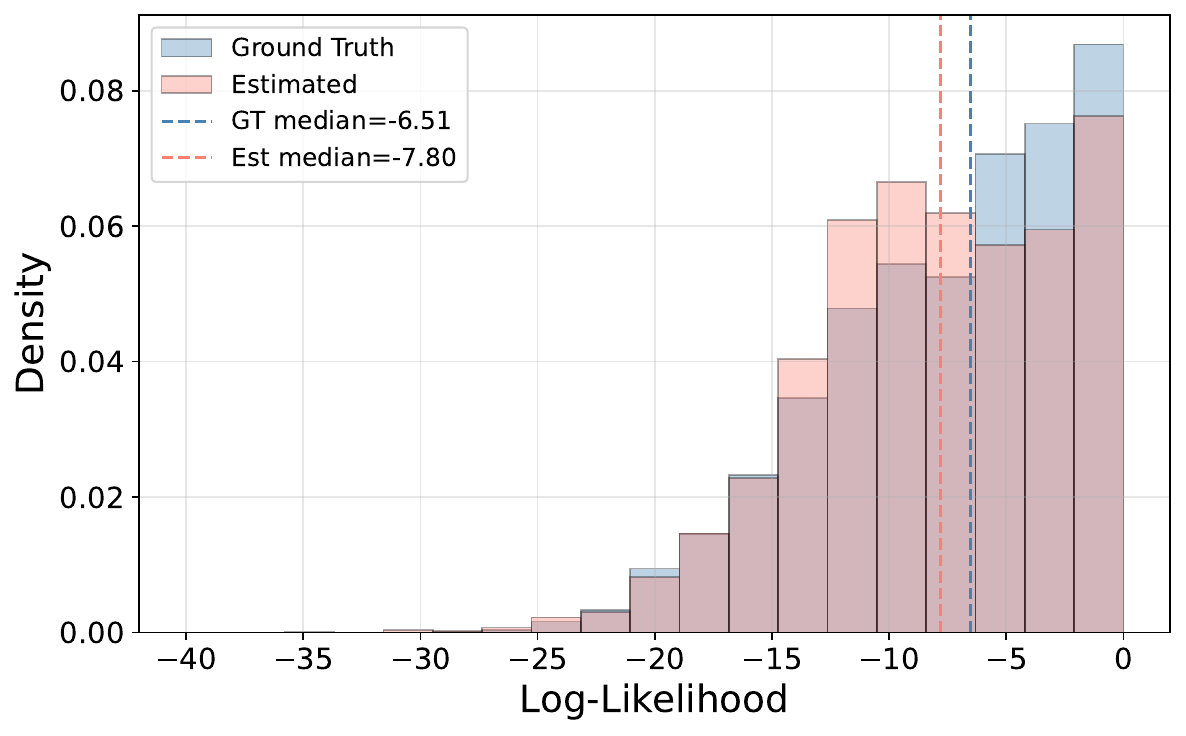}
  \caption{Comparison of trajectory log-likelihood distributions for ground-truth (blue) and estimated (red) trajectories, evaluated under the ground-truth transition matrix.}
  \label{fig:traj_loglik}
\end{figure}

Figure~\ref{fig:traj_loglik} compares the distributions of trajectory log-likelihoods computed for ground-truth and estimated trajectories, both evaluated under the ground-truth time-dependent transition matrix $Q$, this last averaged over all time intervals. 

Both distributions peak in the range $[-5,\, 0]$, which indicates that the majority of estimated trajectories follow high-probability transitions according to the ground-truth model. However, the estimated distribution exhibits a heavier left tail extending to approximately $-35$, while the ground-truth distribution is more concentrated near zero. This indicates that some estimated trajectories are not present (or rarely) in the ground truth, most likely arising from the edge-level assignment step, which occasionally selects shortest routes that cross region boundaries in atypical ways. Also, both distributions assign the highest probability mass to the same interval $[-5,\, 0]$, proving that the proposed method recovers the dominant mobility patterns of the synthetic scenario. The residual mismatch in the tail reflects the under-determination of the estimation problem. 

\section{Discussion and Conclusion}\label{sec:conclusion}

The key contribution of this paper is a hierarchical two-level decomposition of the calibration problem: a regional level, where flows are estimated by gradient-based optimization against aggregated sensor counts, and a edge level, where trajectories are refined by scoring against temporal and volumetric sensor profiles. This decomposition ensures regional concordance (agreement between estimated and observed counts at the region level), a property that global calibration approaches do not enforce. The approach is simulation-free, differentiable, and approximately 180× faster than the simulation-based baseline.

Experiments on the Brussels road network with real data from 133 sensors show that the proposed method reproduces the daily traffic profile with accuracy competitive with a simulation-based baseline~\cite{guastella_calibration_2025} while requiring only approximately 23~seconds for region-level allocation (roughly $180\times$ faster) and outperforming RouteSampler across all reported metrics. Synthetic experiments prove that the estimated transition matrices closely recover the ground-truth mobility structure.

Three limitations deserve attention. First, the estimation problem is under-determined: the weighted contribution matrix and the aggregate consistency term only partially constrain the solution space. In unobserved parts of the network, multiple flow configurations are equally consistent with the data, and there is no principled basis for preferring one over another without additional information such as probe trajectories or prior OD matrices. The proposed method should therefore be understood as recovering a plausible and regionally consistent traffic model, not the true one. Second, edge-level trajectory refinement relies on shortest-route heuristics, which do not capture realistic route choice under congestion. Third, the method is sensitive to the number of optimization iterations: beyond a problem-dependent threshold, the optimizer concentrates flows on sensored routes, degrading accuracy in unobserved corridors.

The under-determined nature of the estimation problem also introduces a risk of overfitting to specific sensor counts. To mitigate this, the solution from time $t$ can be used to initialize the optimization for $t+1$. This acts as a temporal regularizer, assuming that traffic flows are continuous and that drastic changes in route distribution are physically unlikely between consecutive time intervals. Further regularization strategies, such as penalizing route complexity or using spatial proximity constraints, can prevent the model from over-fitting to local sensor noise, ensuring that the results reflect global mobility patterns.

Several future directions are promising. Integrating complementary data sources such as probe trajectories, floating-car data, or mobile network indicators, would further constrain the estimation problem and reduce OD ambiguity. Replacing shortest-route heuristics with learning-based route choice models would improve trajectory realism without requiring full simulation. Finally, meta-learning optimization approaches to hyperparameter tuning would eliminate the current reliance on empirical parameter search.

\bibliographystyle{plain}
\bibliography{biblio}

@article{ENGLEZOU2024213,
  title        = {Dynamic origin-destination matrix estimation for networks operating under free-flow conditions using macroscopic flow dynamics},
  author       = {Englezou, Y. and Timotheou, S. and Panayiotou, C. G.},
  year         = 2024,
  journal      = {IFAC-PapersOnLine},
  volume       = 58,
  number       = 10,
  pages        = {213--218},
  doi          = {10.1016/j.ifacol.2024.07.342},
  issn         = {2405-8963},
  note         = {17th IFAC Symposium on Control of Transportation Systems CTS 2024}
}

@article{GALLIANI2024104246,
  title = {Estimation of dynamic Origin–Destination matrices in a railway transportation network integrating ticket sales and passenger count data},
  journal = {Transportation Research Part A: Policy and Practice},
  volume = {190},
  pages = {104246},
  year = {2024},
  issn = {0965-8564},
  doi = {10.1016/j.tra.2024.104246},
  author = {Galliani, G. and Secchi, P. and Ieva, F.}
}

@article{routesampler,
  title        = {A comparison of SUMO's count based and countless demand generation tools},
  author       = {Behrisch, M. and Hartwig, P.},
  year         = 2022,
  month        = {Jun.},
  journal      = {SUMO Conference Proceedings},
  volume       = 2,
  pages        = {125–131},
  doi          = {10.52825/scp.v2i.107}
}

@ARTICLE{hellinger,
  author={Sadiq, M. and Kadhim, M. N. and Al-Shammary, D. and Milanova, M.},
  journal={IEEE Access}, 
  title={Novel EEG Classification Based on Hellinger Distance for Seizure Epilepsy Detection}, 
  year={2024},
  volume={12},
  pages={127357-127367},
  doi={10.1109/ACCESS.2024.3450449}
}

@article{Bhattacharyya,
  title = {Improving burn diagnosis in medical image retrieval from grafting burn samples using B-coefficients and the CLAHE algorithm},
  journal = {Biomedical Signal Processing and Control},
  volume = {99},
  pages = {106814},
  year = {2025},
  issn = {1746-8094},
  doi = {10.1016/j.bspc.2024.106814},
  author = {Rangaiah, P. K.B. and Pradeep kumar, B.P. and Augustine, R.}
}

@inproceedings{emode23,
  title        = {Traffic Simulation with Incomplete Data: the Case of Brussels},
  author       = {Guastella, D. A. and Cornelis, B. and Bontempi, G.},
  year         = 2023,
  booktitle    = {Proceedings of the 1st ACM SIGSPATIAL International Workshop on Methods for Enriched Mobility Data: Emerging Issues and Ethical Perspectives 2023},
  location     = {Hamburg, Germany},
  publisher    = {Association for Computing Machinery},
  address      = {New York, NY, USA},
  series       = {EMODE '23},
  pages        = {15–24},
  doi          = {10.1145/3615885.3628004},
  isbn         = 9798400703478
}

@article{nguyen_net,
  title        = {An Efficient Method for Computing Traffic Equilibria in Networks with Asymmetric Transportation Costs},
  author       = {Nguyen, S. and Dupuis, C.},
  year         = {1984},
  journal      = {Transportation Science},
  publisher    = {INFORMS},
  volume       = 18,
  number       = 2,
  pages        = {185--202},
  issn         = {00411655, 15265447}
}

@article{tang_parallel-computing-based_2024,
  title        = {Parallel-Computing-Based Calibration for Microscopic Traffic Simulation Model},
  author       = {Tang, L. and Zhang, D. and Han, Y. and Fu, A. and Zhang, H. and Tian, Y. and Yue, L. and Wang, D. and Sun, J.},
  journal      = {Transportation Research Record},
  volume       = 2678,
  number       = 4,
  pages        = {279--294},
  doi          = {10.1177/03611981231184244},
  year         = {2024}
}

@article{daguano_automatic_2023,
  title        = {Automatic Calibration of Microscopic Traffic Simulation Models Using Artificial Neural Networks},
  author       = {Daguano, R. F. and Yoshioka, L. R. and Netto, M. L. and Marte, C. L. and Isler, C. A. and Santos, M. M. D. and Justo, J. F.},
  volume       = 23,
  number       = 21,
  pages        = 8798,
  doi          = {10.3390/s23218798},
  journal      = {Sensors},
  year         = {2023}
}

@article{roocroft_flow_2025,
  title        = {Flow count data-driven static traffic assignment models through network modularity partitioning},
  author       = {Roocroft, A. and Punzo, G. and Ramli, M. A.},
  journal      = {Transportation},
  volume       = 52,
  number       = 1,
  pages        = {185--214},
  doi          = {10.1007/s11116-023-10416-x},
  year         = {2025}
}

@article{CAO2024104850,
  title        = {Data driven origin–destination matrix estimation on large networks-A joint origin-destination-path-choice formulation},
  author       = {Cao, Y. and Van Lint, H. and Krishnakumari, P. and Bliemer, M.},
  year         = 2024,
  journal      = {Transportation Research Part C: Emerging Technologies},
  volume       = 168,
  pages        = {104850},
  doi          = {10.1016/j.trc.2024.104850}
}

@article{WEI2025668,
  title        = {Consistent origin-destination and link flow estimation based on data-driven network assignment},
  author       = {Wei, G. and Gundleg{\aa}rd, D. and Rydergren, C.},
  year         = 2025,
  journal      = {Transportation Research Procedia},
  volume       = 86,
  pages        = {668--675},
  doi          = {10.1016/j.trpro.2025.04.083}
}

@misc{zhang2025osorio,
  title        = {Origin-Destination Travel Demand Estimation: An Approach That Scales Worldwide, and Its Application to Five Metropolitan Highway Networks},
  author       = {Zhang, C. and Arora, N. and Bian, C. and Li, Y. and Ng, W. and Tomkins, A. and Yan, B. and Zhang, J. and Osorio, C.},
  year         = 2025,
  eprint       = {2507.00306},
  archiveprefix = {arXiv},
  primaryclass = {cs.ET}
}

@article{ZHANG2026129002,
  title        = {An improvement by introducing LBFGS idea into the Adam optimizer for machine learning},
  author       = {Zhang, Z. and Yuan, G. and Qin, Z. and Luo, Q.},
  year         = {2026},
  journal      = {Expert Systems with Applications},
  volume       = 296,
  pages        = 129002,
  doi          = {10.1016/j.eswa.2025.129002}
}

@article{guastella_calibration_2025,
  title        = {Calibration of vehicular traffic simulation models by local optimization},
  author       = {Guastella, D. A. and Morales-Hernández, A. and Cornelis, B. and Bontempi, G.},
  journal      = {Transportation},
  year         = {2025},
  doi          = {10.1007/s11116-025-10593-x}
}

@article{OSORIO201918,
    title = {High-dimensional offline origin-destination (OD) demand calibration for stochastic traffic simulators of large-scale road networks},
    journal = {Transportation Research Part B: Methodological},
    volume = {124},
    pages = {18-43},
    year = {2019},
    issn = {0191-2615},
    doi = {10.1016/j.trb.2019.01.005},
    author = {Osorio, C.},
}

\end{document}